\documentclass[10pt,twocolumn,letterpaper]{article}
\pdfoutput = 1
\usepackage{iccv}
\usepackage{times}
\usepackage{epsfig}
\usepackage{graphicx}
\usepackage{amsmath}
\usepackage{amssymb}
\usepackage{multirow}
\usepackage{subfig}
\usepackage{comment}
\usepackage{algorithmic}
\usepackage{enumitem}
\usepackage{mathtools}
\usepackage{amsfonts}

\usepackage[pagebackref=true,letterpaper=true,colorlinks,bookmarks=false]{hyperref}

 \iccvfinalcopy 

\def\iccvPaperID{****} 

\newcommand{\argmin}[1]{\underset{#1}{\operatorname{argmin}}}

\newcommand{\bA}{\mathbf{A}}
\newcommand{\bB}{\mathbf{B}}
\newcommand{\bC}{\mathbf{C}}

\newcommand{\bH}{\mathbf{H}}
\newcommand{\bI}{\mathbf{I}}

\newcommand{\bM}{\mathbf{M}}

\newcommand{\bV}{\mathbf{V}}
\newcommand{\bW}{\mathbf{W}}
\newcommand{\bX}{\mathbf{X}}
\newcommand{\bY}{\mathbf{Y}}

\newcommand{\bc}{\mathbf{c}}

\newcommand{\bx}{\mathbf{x}}
\newcommand{\by}{\mathbf{y}}

\newcommand{\real}{\mathbb{R}}

\newcommand{\order}{\mathcal{O}}

\long\def\ignorethis#1{}

\newsavebox{\savepar}

\newcommand{\trans}[1]{{#1}^{\ensuremath{\mathsf{T}}}}

\graphicspath{{Figs/}}
\newcommand*\samethanks[1][\value{footnote}]{\footnotemark[#1]}
\ificcvfinal\fi
\begin{document}


\title{On Large-Scale Retrieval: Binary or $n$-ary Coding?}
\author{Mahyar Najibi\thanks{Authors contributed equally} \quad \quad \quad Mohammad Rastegari\samethanks \quad \quad \quad Larry S. Davis \\
University of Maryland, College Park\\
{\tt\small\{najibi,mrastega\}@cs.umd.edu \quad \quad \quad lsd@umiacs.umd.edu}}

\maketitle

\begin{abstract}
The growing amount of data available in modern-day datasets makes the need to efficiently search and retrieve information. To make large-scale search feasible, Distance Estimation and Subset Indexing are the main approaches. Although binary coding has been popular for implementing both techniques, $n$-ary coding (known as Product Quantization) is also very effective for Distance Estimation. However, their relative performance has not been studied for Subset Indexing. We investigate whether binary or n-ary coding works better under different retrieval strategies. This leads to the design of a new n-ary coding method,  "Linear Subspace Quantization (LSQ)" which, unlike other $n$-ary encoders, can be used as a similarity-preserving embedding. Experiments on image retrieval show that when Distance Estimation is used, $n$-ary LSQ outperforms other methods. However, when Subset Indexing is applied, interestingly, binary codings are more effective and binary LSQ achieves the best accuracy.

\end{abstract}

\section{Introduction}
Large-scale retrieval has attracted a growing attention in recent years due to the need for image search based on visual content and the availability of large-scale datasets. This paper focuses on the problem of approximate nearest neighbor (ANN) search for large-scale retrieval.

Approaches for solving this problem generally fall into two subcategories; \textbf{Fast Distance Estimation} \cite{jegou2009searching,norouzi2013cartesian} and \textbf{Fast Subset Indexing}\cite{Quinlan:1986,Punjani:2012,LSH,kulis2009kernelized, andoni2006near}.  Fast Distance Estimation methods reduce computation cost by approximating the distance function. Distance computation is very expensive in high dimensional feature spaces. On the other hand, Fast Subset Indexing methods reduce the cost by constraining the search space for a query to a subset of the dataset instead of the whole dataset.

\begin{figure}
\centering
\includegraphics[width=0.45\textwidth,natwidth = 1096, natheight = 907]{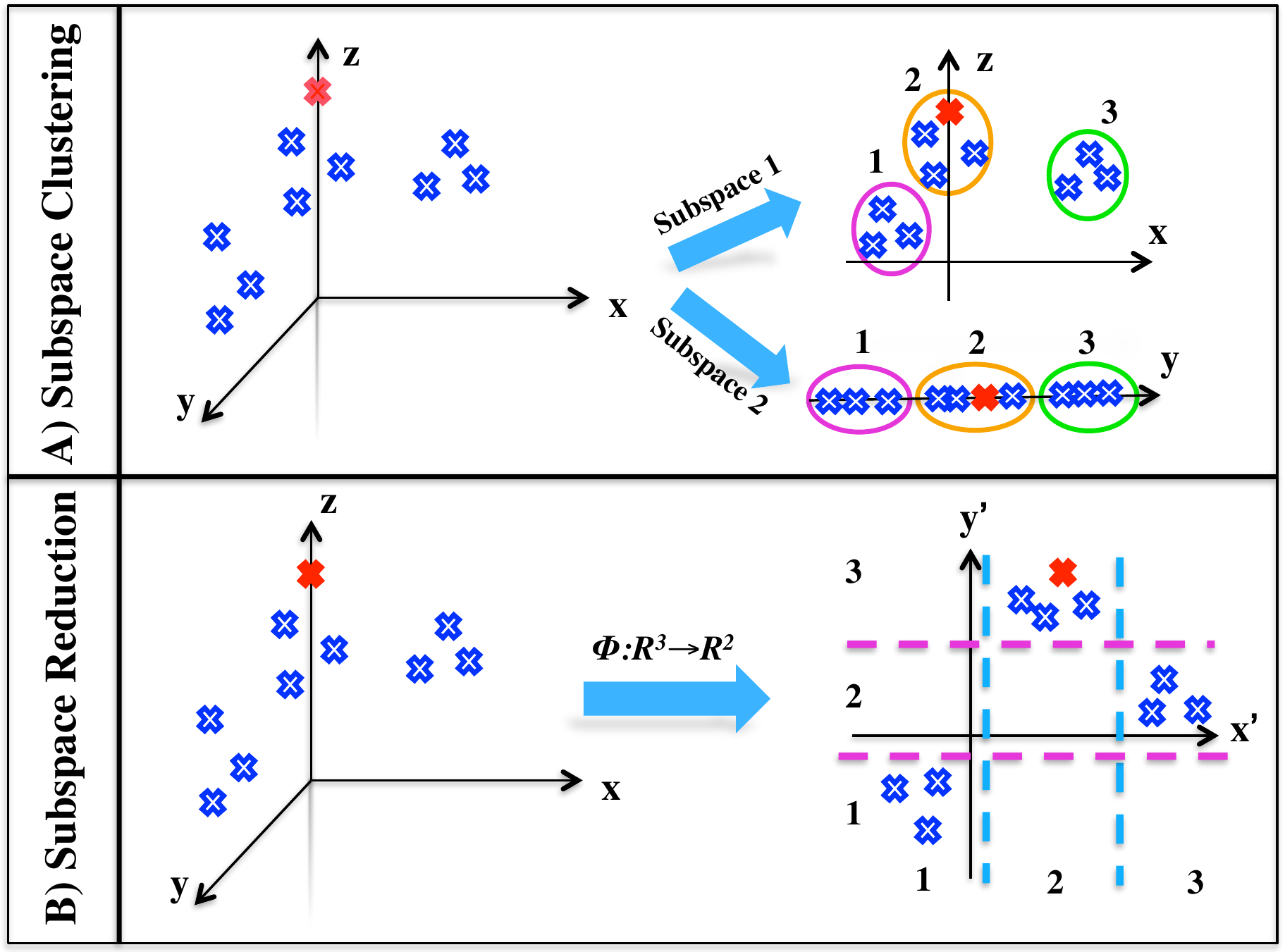}
\caption{\footnotesize Difference between Subspace Clustering (Part A) and Subspace Reduction (Part B) for generating $n$-ary codes. The example shows the simple case of generating 2-dimensional 3-ary codes. In Subspace Clustering, data is clustered into 3 clusters in the two defined subspaces (e.g. the code for the red cross is $[2,2]^T$). In Subspace Reduction, data is transformed into a two dimensional space and each dimension is discretized into 3 bins (e.g. the code for the red cross is $[2,3]^T$).}
\label{fig:figure1}
\end{figure}

A general technique for ANN search (both Fast Distance Estimation and Fast Subset Indexing) is to discritize the feature space into $K$ regions. Different coding methods can be used for this purpose. One of the classic methods is one-hot encoding using $K$-means. $K$-means is a classic quantization technique that quantizes data into $K$ regions (clusters). Data is coded using a $K$ bit binary code, in which only one bit is one (representing the appropriate cluster) . Although $K$-means works well for small values of $K$, it becomes intractable for large $K$. 

An alternative method to one-hot encoding using $K$-means is binary coding. One can code the $K$ clusters with $m = log_2 (K)$\footnote{In this paper, without loss of generality, we assume that $K$ is selected such that $m$ is a natural number.} dimensional binary codes by relaxing the one-hot encoding constraint and allowing multiple bits to be one. This is equivalent to partitioning the space into two regions $m$ times. Many binary coding methods have been designed to address this problem\cite{LSH,ITQ,MLH,DBC}. While binary coding is more scalable, it has a high reconstruction error.

Binary coding can be relaxed by allowing each dimension to be $n$-ary instead of binary (i.e. take on integer values between 1 and $n$). In this case, $K$ clusters can be coded with $m=log_n(K)$ dimensional $n$-ary codes. We introduce a new categorization for methods that generate $n$-ary codes. We explore two general approaches to generate $m$-dimensional $n$-ary codes: 1- \textbf{Subspace Clustering}: In this approach, the original feature space is divided into $m$ subspaces and each subspace is quantized into $n$ clusters. 2- \textbf{Subspace Reduction:} Here, the dimensionality of the original feature space is reduced to $m$ and each dimension is quantized into $n$ bins. Figure \ref{fig:figure1} illustrates these approaches. Multi-dimensional quantization methods (e.g. PQ, CK-means)\cite{ge2014optimized,norouzi2013cartesian,jegou2009searching} adopt the first approach to perform $n$-ary coding. They solve the problem for any $m$ and $n$ (including $n=2$ which leads to binary coding based on the first approach). On the other hand, many binary coding methods(e.g. ITQ, LSH) \cite{ITQ, DBC, LSH,MLH} are instances of the second approach. However they are limited to the case where $n = 2$.

Most recent papers on quantization \cite{ge2014optimized,norouzi2013cartesian}, compared their proposed methods with binary coding methods only with respect to \textit{Distance Estimation} (i.e. typically employing exhaustive search over the data, where the approximated distance is mimicking the ordering of images based on Euclidean distance in the original feature space). This leaves the question of "which binary or $n$-ary coding performs better for ANN search using \textit{Subset Indexing}?" unanswered.

The contributions of this paper are twofold. \textit{First}, a new general approach for multi-dimensional $n$-ary coding is introduced. Based on that, Linear Subspace Quantization (LSQ) is proposed as a new multi-dimensional $n$-ary encoder. Unlike previously proposed $n$-ary encoders in which the Euclidean distance between $n$-ary codes is not preserved, the distances in LSQ coded space correlate with the Euclidean distance in the original space. As a result, the codes can be used directly for learning tasks. Furthermore, LSQ does not make the restrictive assumption of dividing space into independent subspaces, which is common in $n$-ary encoders. Experiments show that LSQ outperforms such encoders. \textit{Second}, it is shown that $n$-ary coding does not always outperform binary coding in retrieval. We show that binary coding works better when Subset Indexing is used and present an explanation based on the two approaches to coding. To the best of our knowledge this has not been identified previously. However, it is very important for large-scale retrieval.   

The rest of paper is organized as follows. In section \ref{sec:nary}, the general formulation for both Subspace Clustering and Subspace Reduction is presented. Additionally, the LSQ coding method is described and its relation to other methods and its properties are discussed. Section \ref{sec:knn_ret}, describes the ways $n$-ary and binary coding methods can be exploited in combination with distance estimation and subset indexing to reduce search cost in retrieval. Experiments are reported in Section \ref{sec:exp}. Finally, Section \ref{sec:con} concludes the paper.


\section{$n$-ary Coding}
\label{sec:nary}
ANN search methods discretize the feature space into a set of disjoint regions. $n$-ary coding can be used for this purpose. An $n$-ary code of length $m$ is defined by an $m$-dimensional vector in $\lbrace1,2,\hdots, n\rbrace^{m}$. The goal is to transform data into $m$-dimensional $n$-ary codes that reconstruct the original data accurately. 

First, consider constructing a \textit{one}-dimensional $n$-ary code. A common objective for quantization methods is to minimize the reconstruction error, referred to as \textit{quantization error}. In other words, given a set of data points $\bX \in \real^{D \times N}$ where each column is a data point $\bx \in \real^{D}$, the quantization objective can be expressed as: 
\begin{eqnarray}
\label{eq:recon_min}
\min_{Q}\lbrace\Vert\bX-Q(\bX)\Vert_F^2\rbrace
\end{eqnarray}
where $Q$ maps a vector $\bx$ (column in $\bX$) into one element of a finite set of vectors $\bC =\{ \bc_1,\bc_2,\hdots \bc_n \}$ in $\real^{D}$ referred to as a codebook. The index of the codebook vector assigned to a data point is its one-dimensional $n$-ary code. $K$-means optimizes this objective when the size of the codebook is equal to $K$.

Using the one-hot encoding notation, the optimization in \ref{eq:recon_min} can be written as follows:
\begin{eqnarray}
\label{eq:recon_min_Cb}
\min_{\bB,\bC}\lbrace\Vert\bX-\bC\bB\Vert_F^2\rbrace
\end{eqnarray}
where $\bB \in \lbrace 0,1\rbrace^{K \times N}$ is a binary matrix in which each column is a $1$-way selector - all of its elements but one are zero.

In order to generalize one-dimensional $n$-ary codes to $m$-dimensional codes, we explore two approaches: \textit{Subspace Clustering} and \textit{Subspace Reduction}. Although the former has been explored in the literature, the latter has not. Without loss of generality, we assume that the data are mean centered (i.e. $mean(\bX)=0_{D\times 1}$) and scaled to $[-1,1]$ by mapping the data to the unit hyper-sphere. In \cite{rastegari2014comparing, ITQ}, it is shown that it is very beneficial to normalize the data to the unit hyper-sphere.           

\subsection{Subspace Clustering}
Here, to generate $m$-dimensional $n$-ary codes, the original feature space is divided into $m$ subspaces and each subspace is discretized into $n$ regions. To this end, in \ref{eq:recon_min_Cb}, the number of clusters can be set to $K=n^m$ and the selector can be allowed to include $m$ non-zero elements as follows:

\begin{eqnarray}
\label{eq:recon_min_Cb_mdim}
\min_{\bB,\bC}\lbrace\Vert\bX-
\left[\begin{array}{c | c | c | c}
\bC_1 & \bC_2 & \hdots & \bC_m
\end{array}\right]
\left[\begin{array}{c}
\bB_1\\\hline
\bB_2\\\hline
\vdots\\\hline
\bB_m
\end{array}\right]
\Vert_F^2\rbrace
\end{eqnarray}
   
Here $\bC_i$ and $\bB_i$ are the codebook and its related one-hot encoding in the $i$'th subspace. In general, the optimization of \ref{eq:recon_min_Cb_mdim} is intractable. As a result, Product Quantization\cite{jegou2009searching} and Cartesian K-means \cite{norouzi2013cartesian} solve a constrained version of this problem where the subspaces created by the $\bC_i$s are orthogonal. In other words, $\{\forall i,j | ~i\neq j~~\trans{\bC_i}\bC_j=0_{n\times n}$\}. Intuitively, the original space is divided into $m$ independent subspaces and each is clustered into $n$ regions. We next present another approach to $n$-ary coding in which no such constraint is imposed.        

\subsection{Subspace Reduction}
Subspace Reduction maps the data into an $m$-dimensional space and discretizes each dimension into $n$ bins. The goal is to perform this discretization to minimize the reconstruction error in the original space. Formally, the optimization problem can be written as follows: 
\begin{eqnarray}
\label{eq:recon_min_sub}
\min_{f,\tilde{f}}\lbrace\Vert\bX-\tilde{f}(q_n(f(\bX)))\Vert_F^2+\lambda R(\tilde{f})\rbrace
\end{eqnarray}
where $f: \real^{D}\mapsto \real^{m}$ is the \textit{mapping function} and is applied to each column of $\bX$,  $\tilde{f}: \real^{m} \mapsto \real^{D},~(m\leq D)$ is the \textit{reconstruction function} which projects the data back to the $D$ dimensional space in which the reconstruction error is computed. In order to prevent overfitting, the reconstruction function must be regularized. $R$ is a regularizing function that limits the variations in $\tilde{f}$, $\lambda$ is a parameter controlling the amount of regularization, and $q_n$ is a uniform quantizer that is applied to each element of its input and is defined as:
\begin{eqnarray}
\label{eq:quantizer}
q_n(x) = \left
\{\begin{array}{ll}
\theta_n(1) & ~~~~~~~~~~~~~~~~~~~~~~~~~~~~~x<\frac{\theta_n(1)+\theta_n(2)}{2}\\
\theta_n(2) & ~~~~~\frac{\theta_n(1)+\theta_n(2)}{2} \leq x <\frac{\theta_n(2)+\theta_n(3)}{2}\\ 
\vdots\\
\theta_n(n)  & \frac{\theta_n(n-1)+\theta_n(n)}{2} \leq x 
\end{array}
\right.
\end{eqnarray}
where $\theta_n(i)=-1+\frac{2(i-1)}{n-1}$ generates $n$ uniformly distributed values in $[-1,1]$. In other words, $q_n(x)$ is a general quantizer that maps any real value in [-1,1] into one of $n$ uniformly distributed values in $[-1,1]$.  For example, $q_2(x)$ is the sign function $\lbrace\forall x\geq 0|~sign(x)=1,~~\forall x< 0|~sign(x)=-1\rbrace $  and $q_3(x)$ maps $x$ into one of the three values $\lbrace -1,0,1\rbrace$. 

To summarize, optimizing \ref{eq:recon_min_sub} identifies a mapping and a reconstruction function such that the quantized data in the space generated by the mapping function can be reconstructed accurately by the reconstruction function. It should be noted that an $m$-dimensional $n$-ary code is generated by $q_n(f(\bX))$.  

\subsubsection{Linear Subspace Quantization (LSQ)}
LSQ is a multidimensional $n$-ary coding method based on Subspace Reduction where linear functions are used as the mapping and the reconstruction functions in \ref{eq:recon_min_sub}. Assume that $f(\bx) = \trans{\bW}\bx$ and $\tilde{f}(\bx) = \trans{\bV}\bx$ where $\bW\in \real^{D\times m}$ and $\bV\in \real^{m\times D}$. Employing the Frobenius norm as the regularizing function, the optimization problem in \ref{eq:recon_min_sub} becomes:
\begin{eqnarray}
\label{eq:recon_min_sub_lin}
\min_{\bW,\bV}\lbrace\Vert\bX-\trans{\bV}q_n(\trans{\bW}\bX)\Vert_F^2+\lambda \Vert \bV \Vert_F^2\rbrace
\end{eqnarray}

To solve this problem, we propose a two step iterative algorithm. Subsequently, the convergence of the proposed algorithm will be proven.  

\vspace{0.5em}
\noindent $\bullet$ \textbf{Learning LSQ:}
\vspace{0.5em}

The optimization for $\bW$ and $\bV$ in \ref{eq:recon_min_sub_lin} can be solved by a two step iterative optimization algorithm (i.e. fixing one variable and updating the other). The steps are as follows:

\begin{enumerate}[leftmargin=*]
\item \textbf{\textit{Fix $\bW$ and update $\bV$:}} For a fixed $\bW$, define $\bH=q_n(\trans{\bW}\bX)$; then we have a closed form solution for $\bV$ as 
\begin{eqnarray}
\bV=(\bH\trans{\bH}+\lambda \bI)^{-1}\bH\trans{\bX}
\end{eqnarray}
\item \textbf{\textit{Fix $\bV$ and update $\bW$:}} In this step, $\bW$ is updated as:
\begin{eqnarray}
 \bW = \bV^\dagger
\end{eqnarray}
where $\bV^\dagger$ is the  Moore Penrose pseudoinverse of $\bV$. In the following we prove that the pseudoinverse is an optimal solution for \ref{eq:recon_min_sub_lin} when $\bV$ is fixed.
\end{enumerate}
The algorithm iterates between step 1 and 2 until there is no progress in minimizing \ref{eq:recon_min_sub_lin}. 

\vspace{0.5em}
\noindent $\bullet$ \textbf{Convergence of LSQ:}
\vspace{0.5em}

\noindent In order to prove the convergence of the algorithm, we show that both steps reduce the objective value. The optimality of the first step can be easily shown by simple linear algebra. Here, we focus on proving that the second step reduces the objective value. 

Given that $\argmin{\bY}\lbrace\Vert\bB-\bA\bY\Vert\rbrace = \argmin{\bY}\lbrace\Vert\bA^\dagger\bB-\bY\Vert\rbrace$, the solution of the optimization in \ref{eq:recon_min_sub_lin} for fixed $\bV$ is equivalent to the solution of the following problem:
\begin{eqnarray}
\label{eq:recon_min_sub_lin_alter}
\min_{\bW}\lbrace\Vert\trans{{\bV}^\dagger}\bX-q_n(\trans{\bW}\bX)\Vert_F^2\rbrace
\end{eqnarray}

Defining $\bM=\trans{{\bV}^\dagger}\bX$, the optimal solution for $\bW$ can be formulated as:

\begin{eqnarray}
\begin{aligned}
 \bW^*=&\argmin{\bW}\lbrace\Vert\bM-q_n(\trans{\bW}\bX)\Vert_F^2\rbrace\\
\end{aligned}
\label{eq:W_opt}
\end{eqnarray}
It should be noted that the optimal solution is not unique. Therefore, $\mathcal{W}$ is defined as the optimal solution set for $\bW^*$. The goal is to prove that $\bV^\dagger \in \mathcal{W}$. 

Let $\bA^* = \trans{\bW^*}\bX$. We first prove that $q_n(\bA^*) = q_n(\bM)$. Suppose, to the contrary, that $q_n(\bA^*) \neq q_n(\bM)$. Consequently, there should be at least one $i$ and $j$, such that $q_n(A^*(i,j)) \neq q_n(M(i,j))$. Since $q_n(\bA^*)$ is defined in the optimal solution of the optimization \ref{eq:W_opt}, its corresponding objective value should be less than that of any other feasible point. 
 This leads to the conclusion that $(M(i,j)-q_n(A^*(i,j)))^2<(M(i,j)-q_n(M(i,j)))^2$ (Note that even if more than one element differs between $q_n(\bA^*)$ and $q_n(\bM)$, the inequality holds for at least one of them). However, this contradicts the definition of $q_n(x)$ in \ref{eq:quantizer} since $q_n(x)$ should map $M(i,j)$ into $q_n(A^*(i,j))$ (It should be noted that $q_n(A^*(i,j))$ is in the range of $q_n(M(i,j))$). So for any $i$ and $j$, $q_n(A^*(i,j)) = q_n(M(i,j))$. Therefore, $q_n(\bM)=q_n(\bA^*)$. Considering the definition of $\bM$ and $\bA^*$ completes the proof that $\bV^\dagger \in \mathcal{W}$.
 
  Finally, since both steps in our optimization reduce the objective value, LSQ converges to a local optimal value of optimization \ref{eq:recon_min_sub_lin}.

\vspace{0.5em}
\noindent $\bullet$ \textbf{Relation to ITQ:}
\vspace{0.5em}

\noindent ITQ is a special case of LSQ when $n=2$ and $\bW$, $\bV$ are rotation matrixes where $\bW$=$\trans{\bV}$. Our experiments show that the binary codes generated by LSQ leads to higher accuracies than the binary codes generated by ITQ.
 
\begin{figure}[t]
\centering
\includegraphics[scale=0.4,natwidth = 1203, natheight = 543]{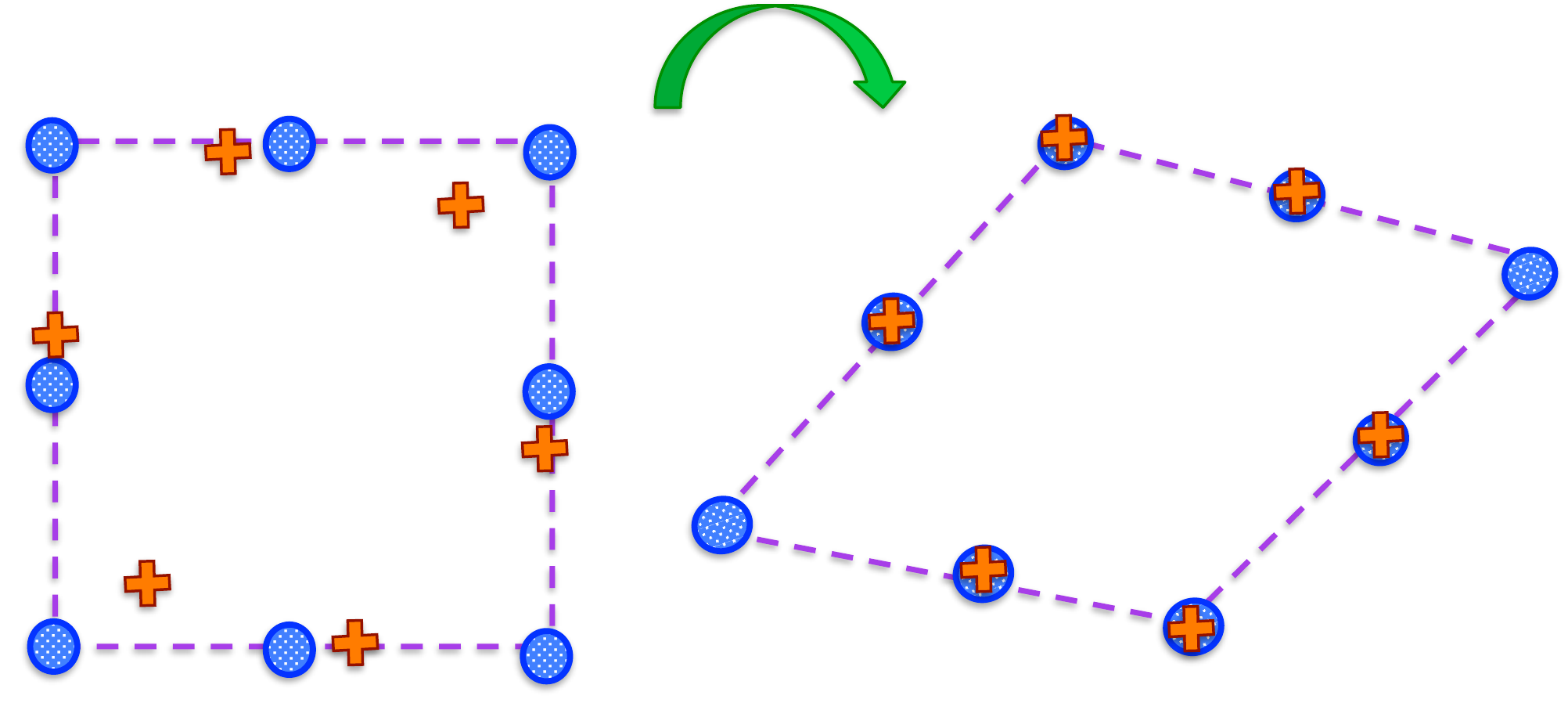} 
\caption{\footnotesize LSQ fitting 2D data using a $q_3$ quantizer. Data points are shown by orange crosses. The blue circles indicate the quantization levels in different dimensions. LSQ reduces the reconstruction error by performing a linear transformation over the quantized hyper-cube.}
\label{fig:lsq_q3}
\end{figure}

\vspace{0.5em}
\noindent $\bullet$ \textbf{Geometrical Interpretation:}
\vspace{0.5em}

\noindent LSQ finds a linear transformation of a quantized hyper-cube that best fits the data. Figure \ref{fig:lsq_q3} illustrates a simple 2D example in which the $q_3$ quantizer is fit to the data by a rotation. 

\subsubsection{LSQ as an $n$-ary Embedding}
\label{sec:embedding}
While binary encoding techniques try to minimize the reconstruction error, the resulting codes preserve similarities between samples. In other words, the Hamming distance in the binary space approximates the Euclidean distance in the original feature space. As a consequence of this property, these binary codes can be exploited as feature vectors for learning tasks in the embedded space. Many recent approaches based on this property have been proposed to make learning more efficient \cite{DBC,torresani2010efficient}.

In subspace clustering methods (e.g. CK-means), the cluster indices generated by the quantizer can not be viewed as a similarity preserving embedding. This is due to the fact that there are no constraints on assigning these indices to clusters. In subspace reduction methods (e.g. LSQ), each dimension of an $n$-ary code has a finite(discrete) set of real values as its domain. For each dimension, the distances between these discrete values correlates with the distances between the data points in the original feature space in the direction of that dimension. Therefore, the Euclidean distance in the quantized data correlates with the Euclidean distance in the original feature space.   

One could post process CK-means to generate similarity(distance) preserving codes by assigning the appropriate indices to cluster centers after completion of the training stage. These indices can be obtained by finding a 1D subspace for each of the subspaces generated by CK-means. A simple model could compute PCA over the cluster centers in each subspace to reduce the cluster centers into 1D real values.  However, in \ref{sec:exp_embedding}, a classification experiment is performed in which the $n$-ary codes are used as features. The result shows that, as an embedding, LSQ outperforms CK-Means by a large margin even after refining the CK-Means index assignments to clusters.

\section{$K$-NN Retrieval using Data Encoding}
\label{sec:knn_ret}
A large source of computational cost in nearest neighbor search is the distance computation between the query and all the samples in the dataset. In order to speed up $K$-NN search, one can either speed up the computation of the distance function and/or reduce the number of distance computations by limiting the search space for a given query. We refer to the first strategy as \textit{Distance Estimation} and the second as \textit{Subset Indexing}. In the following subsections, we show how Subspace Clustering and Subspace Reduction coding techniques can be used for each of these strategies.

\subsection{Retrieval by Distance Estimation}
Data coding can reduce the cost of distance computation since the Euclidean distance can be efficiently estimated in the coding space.

\vspace{0.5em}
\noindent $\bullet$ \textbf{Distance estimation using Subspace Clustering $n$-ary codes:}
\vspace{0.5em}

\noindent Once data is coded, the Euclidean distance between two points can be estimated as the sum of distances between the assigned cluster centers to those data points in each subspace \cite{jegou2009searching}. This is known as the \textit{symmetric distance}. The distances between the $n$ cluster centers in each subspace can be pre-computed in an $n \times n$ table. Then, computing the symmetric distance can be implemented efficiently by $m$ look-ups and additions of table elements, one for each subspace. More formally,
\begin{eqnarray}
\label{eq:dist_quant}
d(\bx,\by)=\sum_{i=1}^{m}L_i(c(u_i(\bx)),c(u_i(\by)))
\end{eqnarray}      
where $u_i(\bx)$ project $\bx$ into the $i^{th}$ subspace, $c(u)$ is the cluster index to which $u$ belongs, and $L_i$ is the pre-computed distance table for the $i^{th}$ subspace. If we consider $\bx$ as the query and $\by$ as a data point from the database, $c(u_i(\by)))$ can be pre-computed.  Therefore the complexity is $\order(mN)$ for each query, where $N$ is the total number of points in the database. 

\vspace{0.5em}
\noindent $\bullet$ \textbf{Distance estimation using subspace reduction $n$-ary codes:}
\vspace{0.5em}

As mentioned earlier, the Euclidean distance between quantized data by subspace reduction approximates the Euclidean distance in the original feature space. Therefore we need only compute the distance between coded values. This has complexity $\order(mN)$, which is the same as the complexity of subspace clustering.  

\vspace{0.5em}
\noindent $\bullet$ \textbf{Distance estimation using binary codes:}
\vspace{0.5em}

\noindent For the binary codes, Hamming distance is used as the distance metric. Computing Hamming distances using $m$-bit binary codes has complexity $\order(mN)$ for each query.

\subsection{Retrieval by Subset Indexing}
\label{sec:ret_subset}

\begin{figure}[t]
\centering
\subfloat[]{\includegraphics[height=0.23\textwidth,natwidth = 4.19in, natheight= 4.11in]{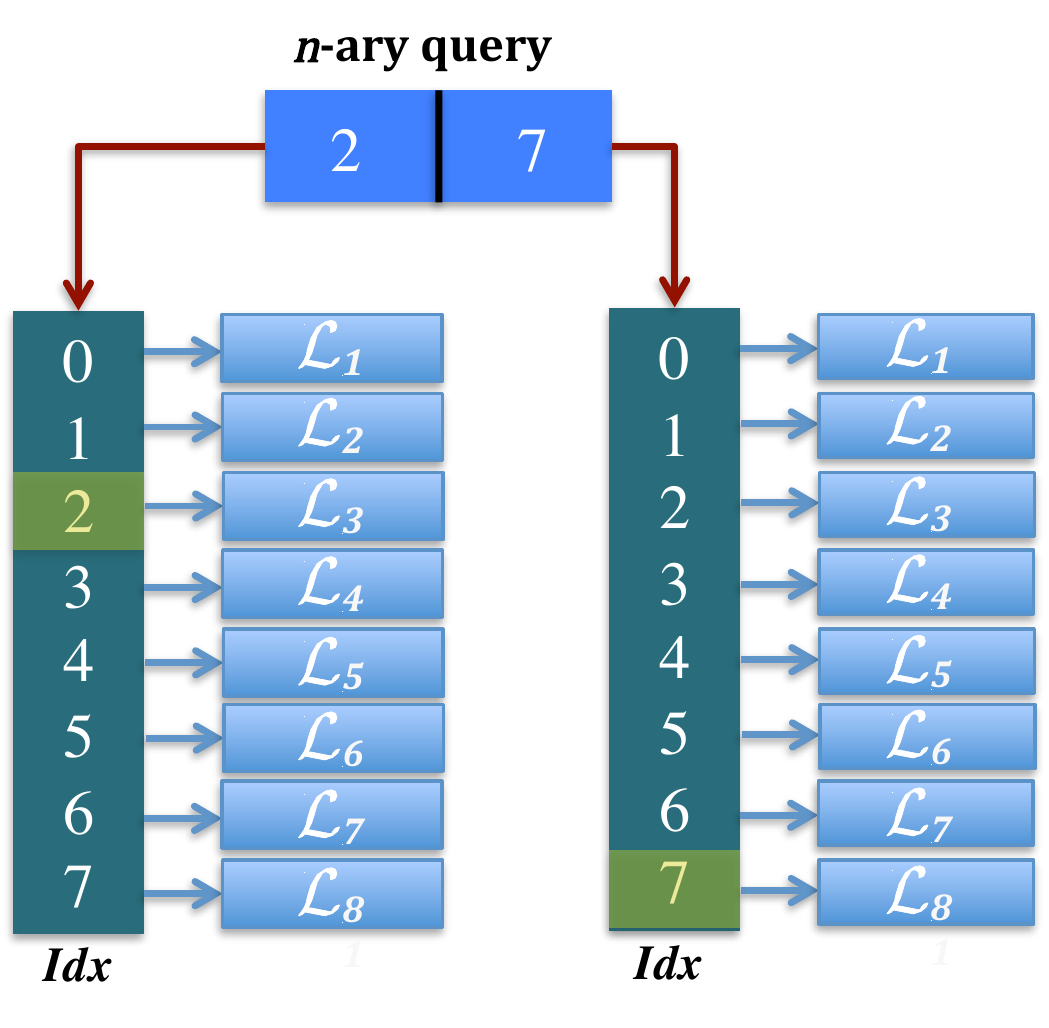}}
\subfloat[]{\includegraphics[height=0.23\textwidth, natwidth = 4.23in, natheight = 4.08in]{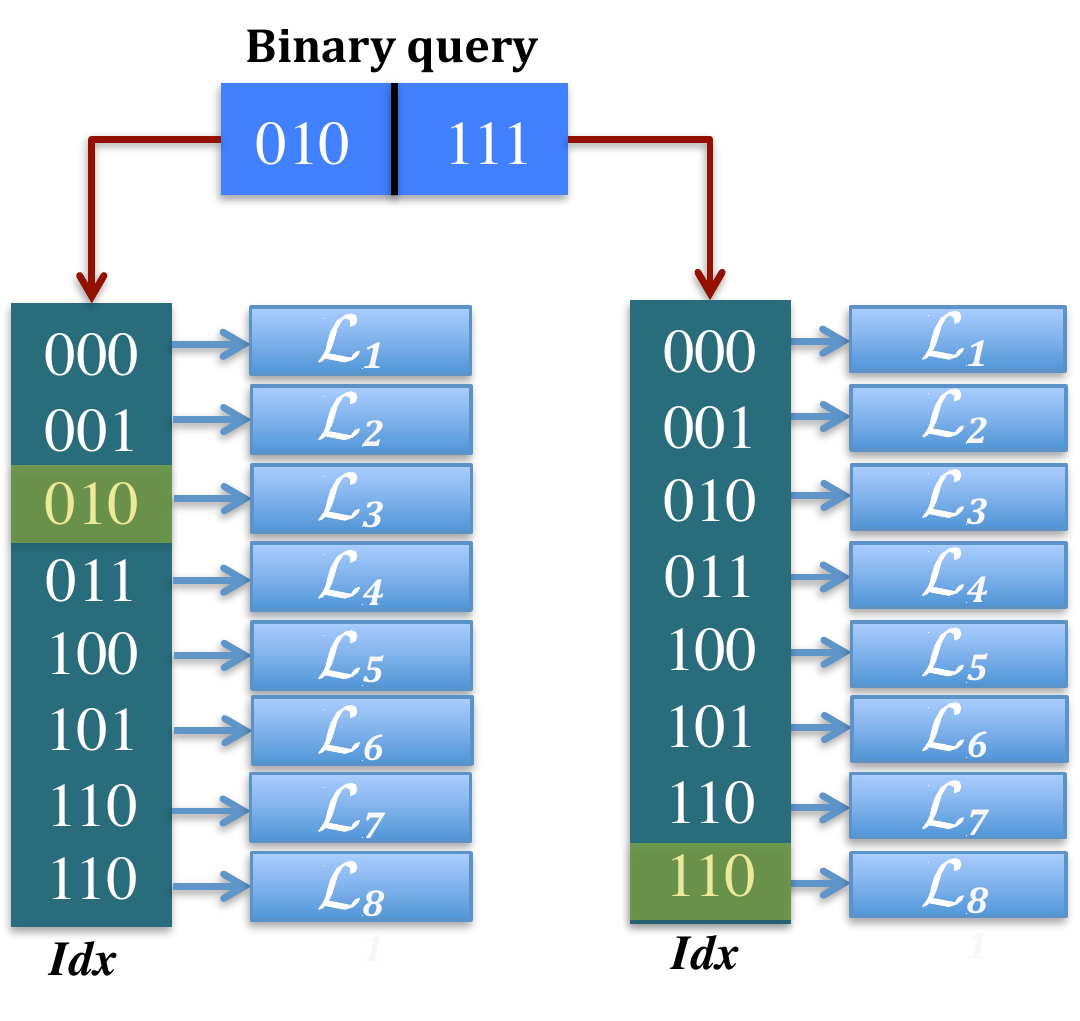}}
\caption{\footnotesize Retrieval using Multiple Index Hashing (a) $n$-ary Coding: each dimension of the query is a key index to its corresponding table. Each table has $n$ rows which points to set of samples in the base set with the same value in that dimension. (b) Binary Coding: In this case, binary codes are partitioned into sets of $b$ consecutive bits (here $b=3$) and each set is used as a key index for the corresponding table. Tables have $2^b$ rows containing samples with the same value in the corresponding bits.}
\label{fig: mult_idx_hash}
\end{figure}
Another way to speed up nearest neighbor search is to limit the search space. Hashing techniques \cite{LSH} and tree based methods\cite{Quinlan:1986} limit the search space by constraining search to a subset of samples in the database. This is accomplished by indexing the data into a data structure (e.g. hash tables or search tree) at training time. Multiple index hashing \cite{greene1994multi,Punjani:2012} is one such data structure that can be used for binary and $n$-ary codes. 
     
\vspace{0.5em}
\noindent $\bullet$ \textbf{Multiple index hashing using $n$-ary codes:}
\vspace{0.5em}     
     
\noindent In this approach, for $m$-dimensional $n$-ary coding, we create an index table $\mathcal{T}_i, ~i=1,\hdots,m$ for each dimension. Each table has $n$ tuples  $(Idx_j, \mathcal{L}_j), ~j=1,\hdots,n.$ where $Idx_j$ corresponds to one of the $n$ values in a dimension of the code and $\mathcal{L}_j$ is a list of those data points' indices such that the value of the $i^{th}$ dimension in their code is $Idx_j$. At query time, for each dimension of the code, a set of data indices is retrieved. Figure \ref{fig: mult_idx_hash}(a) illustrates this technique. For each index in the union of these sets, we assign a score $s$ between $1$ and $m$ which indicates that a particular index has been retrieved from $s$ dimensions. The samples with higher scores are more likely to be similar to the query sample. By sorting the indices based on their score, we can choose the top-$K$ samples as the $K$-NN's. If the total number of retrieved indices were less than $K$, we change the value in one of dimensions in the query code that has minimum distance to the quantized query point in the original space. Then we retrieve a new set and repeat the process until the total size of the retrieval set is greater than or equal to $K$.               

\vspace{0.5em}
\noindent $\bullet$ \textbf{Multiple index hashing with binary codes:}
\vspace{0.5em} 

\noindent Similar to $n$-ary codes, binary codes can be used for multiple index hashing. However, in this case each set of $b$ consecutive bits are grouped together to create the indices for accessing the tables. Considering $b = log(n)$, this partitioned binary code can be seen as an $n$-ary code. As a result, the same technique can be applied for multiple index hashing as discussed previously. This case can be seen in Figure \ref{fig: mult_idx_hash}(b).

\subsubsection{Subset Indexing: Binary or $n$-ary Coding?}
As mentioned earlier, $n$-ary coding does not always outperforms binary coding for large-scale retrieval. More precisely, when Subset Indexing is used to reduce the search cost, binary coding achieves better search accuracy. This is due to the fact that quantization does not necessarily preserve the similarities (or distances) between data. In other words, a good quantizer $Q$ that minimizes the quantization error in \ref{eq:recon_min}, does not always preserve relative distances between data. Formally:
\begin{eqnarray}
\label{eq:quant_ieq}
\begin{aligned}
\Vert\bx_1-\bx_2\Vert\leq\Vert \bx_1-\bx_3\Vert \nRightarrow ~~~~~~~\\
~~~~\Vert Q(\bx_1)-Q(\bx_2)\Vert\leq\Vert Q(\bx_1)-Q(\bx_3)\Vert
\end{aligned}
\end{eqnarray}  

\begin{figure}[t]
\centering
\includegraphics[scale=0.45,natwidth=624,natheight=605]{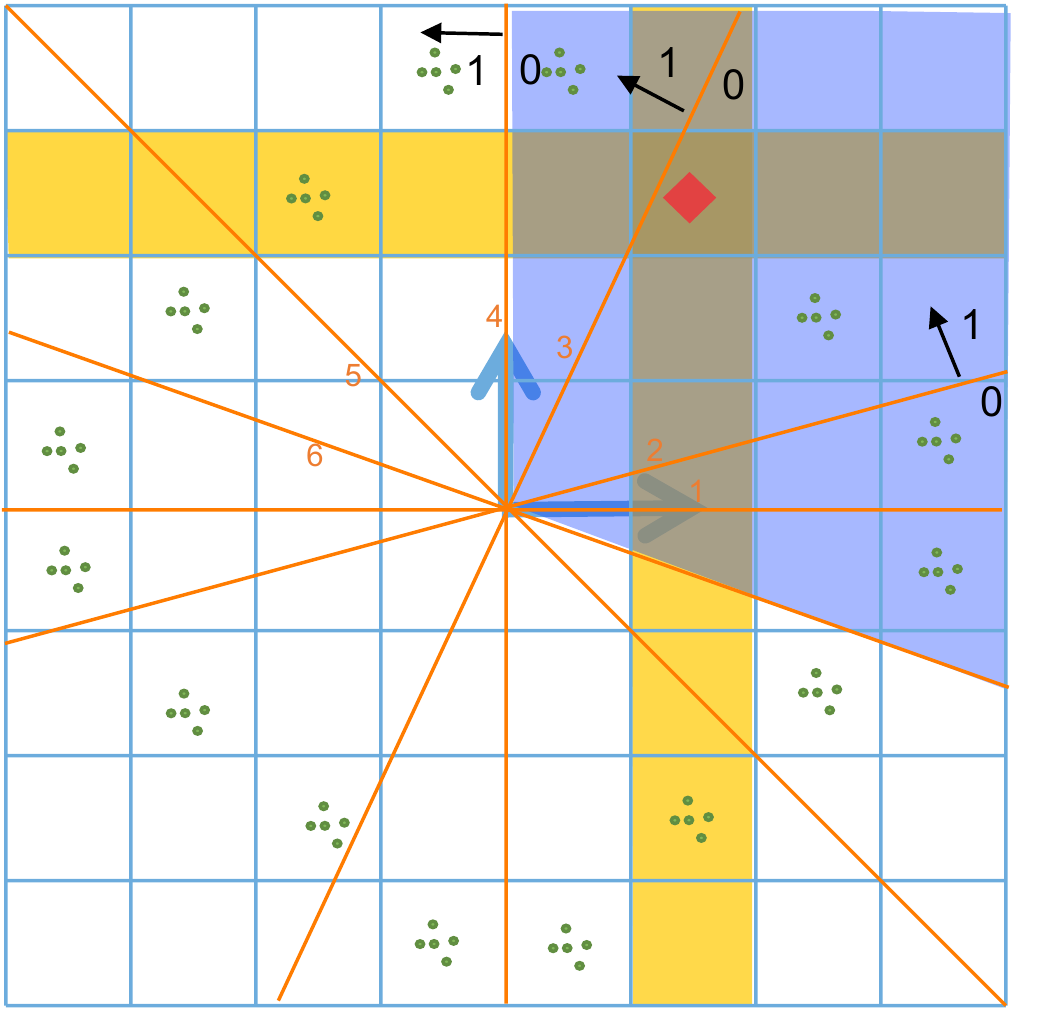} 
\caption{Binary vs. $n$-ary for ANN}
\label{fig: binary_vs_nary}
\end{figure}
This is important when retrieval is carried out by subset indexing. There, binary codes may retrieve the nearest neighbors better than $n$-ary codes. Figure \ref{fig: binary_vs_nary} illustrate an example of $2$-dimensional $8$-ary codes and their corresponding binary codes, which have $6$ bits ($6=2\log(8)$). Each bit is generated by a line based on which side of the line the point lies. The green dots are the points in the database and the red diamond is a query point. In this figure the binary code for the query sample is $110000$. In the subspace clustering view, we cluster each dimension into 8 clusters. In this case, all points in the yellow region will be retrieved by multiple index hashing.  As can be seen, none of the actual nearest neighbors can be retrieved. But, when we use the binary codes for multiple index hashing all the actual nearest neighbors are retrieved. This is the blue region (i.e. the union of the region created by the first three bits ($110$) and the second three bits ($000$) of the query code). Our experimental evaluation confirms that when subset indexing is used for retrieval, binary codes outperform $n$-ary codes. Although, $n$-ary codes are more accurate for quantization, they are not accurate for ANN with subset indexing.


\section{Experiments}
\label{sec:exp}
We report experiments on three well-known datasets, namely \textit{GIST1M} \cite{jegou2009searching}, \textit{CIFAR10} \cite{krizhevsky2009learning}, and a subset of \textit{ImageNet} \cite{deng2009imagenet} which is used for the ILSVRC2012 challenge. GIST1M contains 1M base feature vectors, 500K training samples and 1K query samples. For CIFAR10, we randomly selected 20K samples as our training set, 500 samples as query images and the remaining 39500 images as the base samples. Raw pixel values are used as features for this dataset. The ImageNet ILSVRC2012 dataset consists of 500K training samples,  250K base samples, and 1K query images. We used ConvNet as the state-of-the-art feature extractor for this dataset. The ConvNet features are extracted by Caffe \cite{caffe}. 

Following \cite{norouzi2013cartesian}, we used recall as the performance measure for retrieval. The training set is used to train the coding model and the learned model is applied for coding the base and query set. For each point in the query set, we find its $R$ nearest neighbors and report the recall at $R$. By varying $R$ we draw the recall curves. 

As mentioned earlier, retrieval can be made faster using two approaches: distance estimation and subset indexing. The performance of different methods can vary with respect to which approach is used. Therefore, each method is examined with respect to both and an analysis is presented. The nearest neighbors in the original feature space are defined as ground truth for each query image. For making the comparison fair, in each experiment the number of bits which can be used by each coding method is limited to the same fixed budget. \textit{e.g.} a $2$ dimensional $8$-ary code requires $6$ bits of memory. ($3$ bits per code dimension).  
\subsection{Retrieval using Distance Estimation} 
\begin{figure*}
\centering
\subfloat[]{\includegraphics[height=0.26\textwidth]{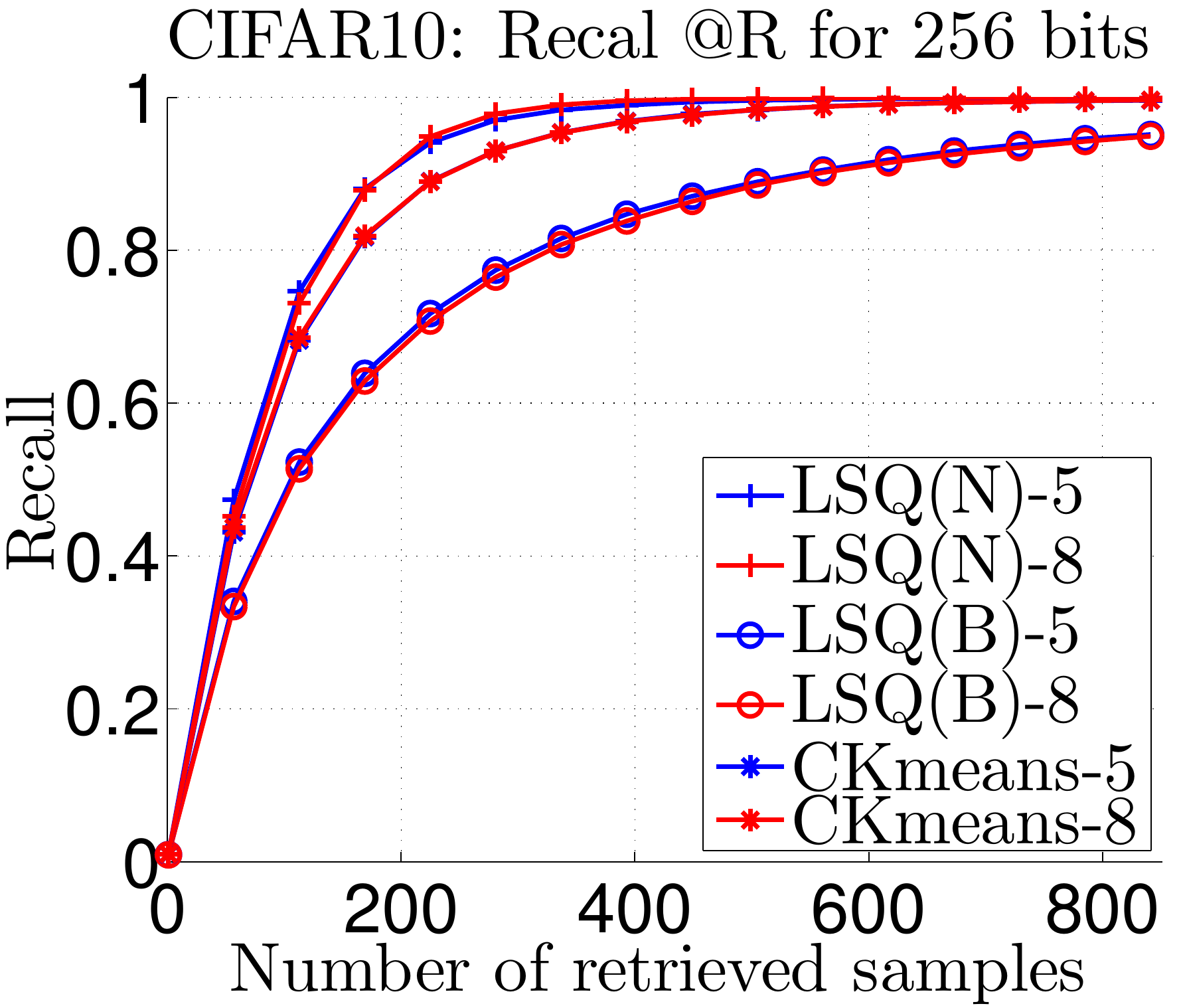}}
\subfloat[]{\includegraphics[height=0.26\textwidth]{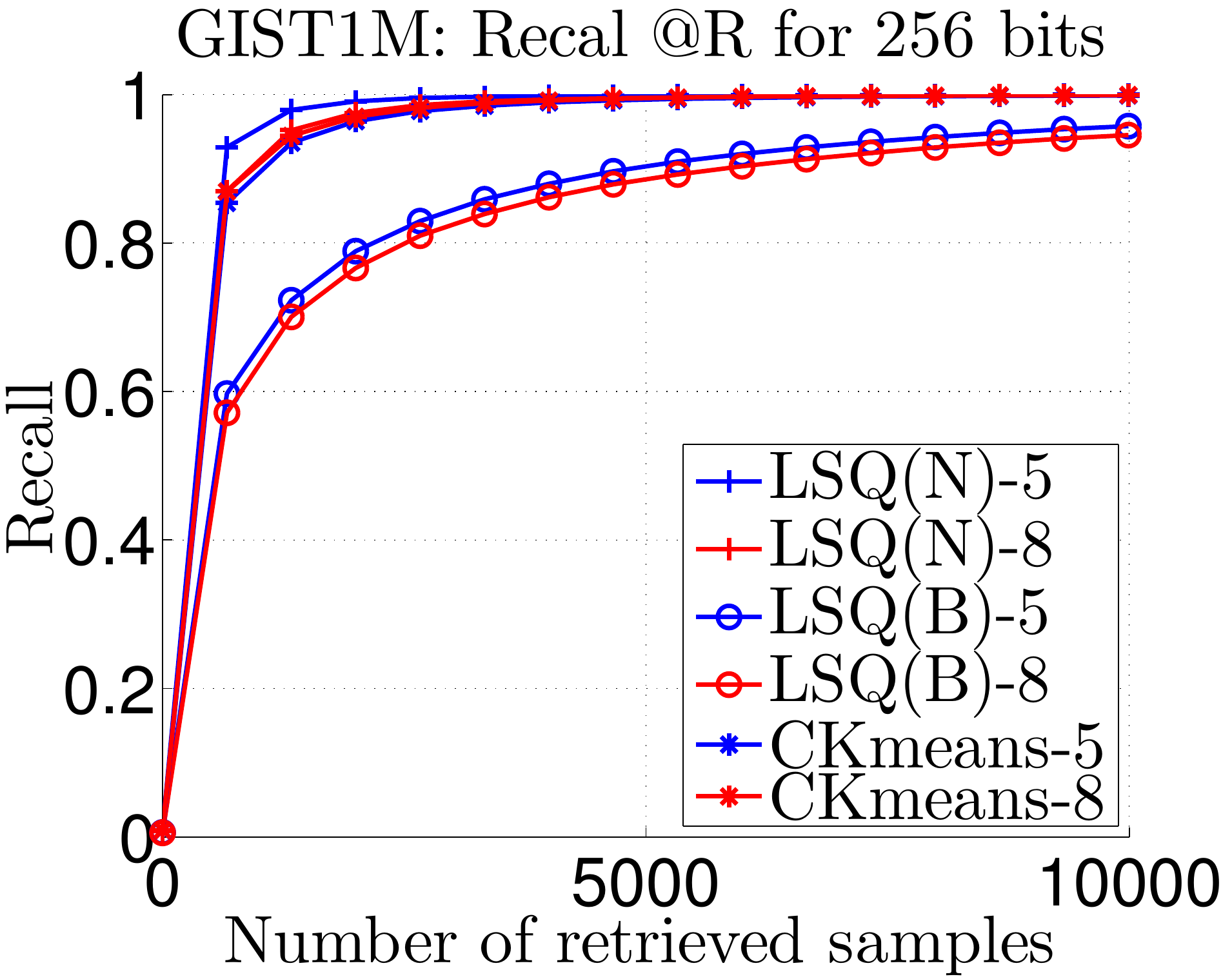}}
\subfloat[]{\includegraphics[height=0.26\textwidth]{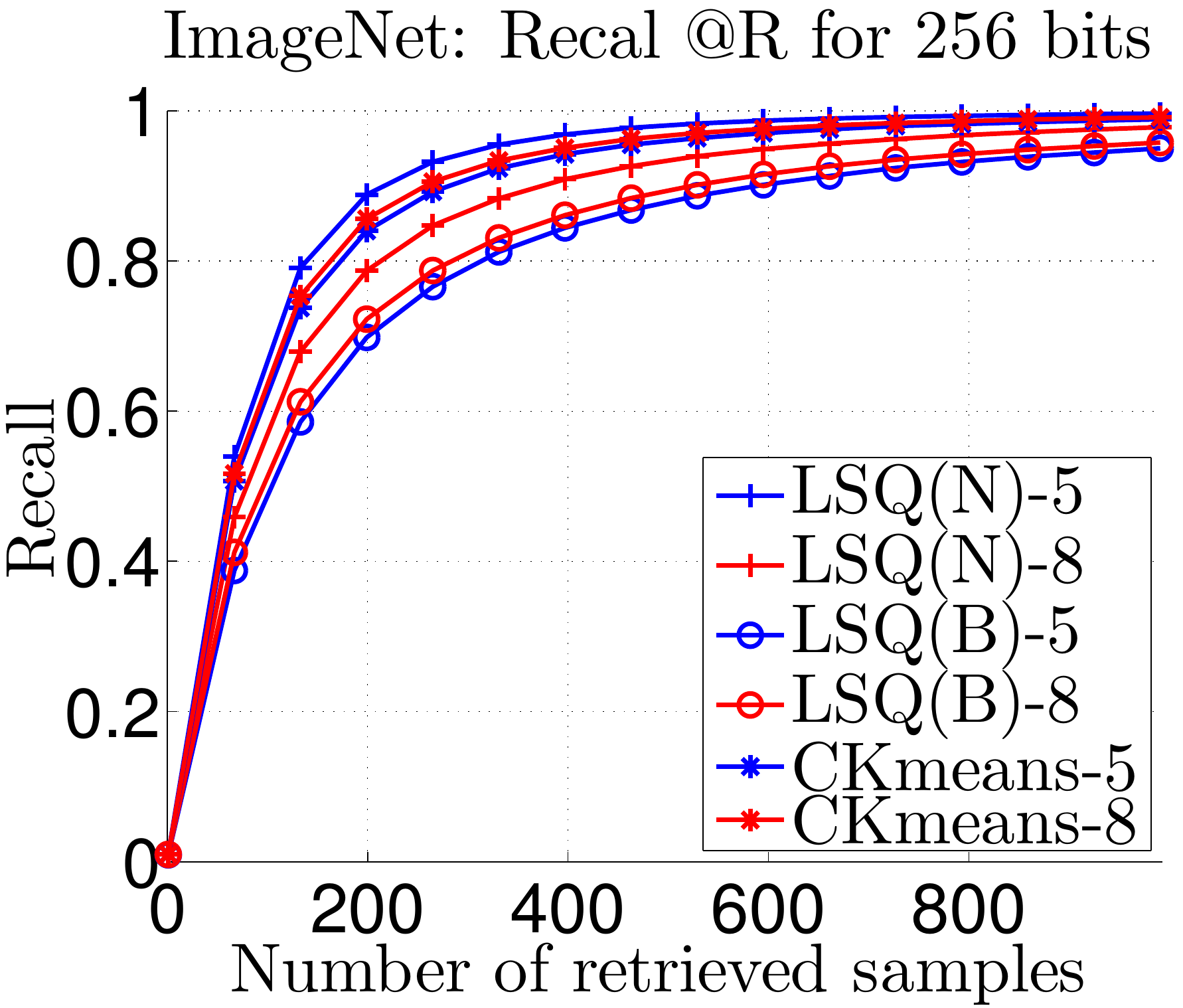}}
\caption{\footnotesize The Recall@R curves for retrieval using distance estimation for 256 bits. Each diagram shows the curve for different methods and different numbers of bits per code dimension. (a) Results on CIFAR10 dataset. (b) Results  on GIST1M dataset. (c) Results on ImageNet dataset.}
\label{fig:distance_estimation}
\end{figure*}

Figure \ref{fig:distance_estimation}, shows the Recall@R curves on different datasets using a budget of 256 bits. In this figure, LSQ(N) and LSQ(B) refer to the n-ary and binary versions of the LSQ method respectively. The recall@R curves are shown for different number of bits per code dimension, which controls the number of quantization steps for n-ary encoders (e.g. for LSQ(N)-5 or LSQ(B)-5 the quantizer has $32$ levels or $5$ bits). As can be seen, the performance of $n$-ary codes is better than binary codes. Also, LSQ outperforms CK-means on all three datasets.

\begin{figure}
\centering
\subfloat[]{\includegraphics[width=0.25\textwidth]{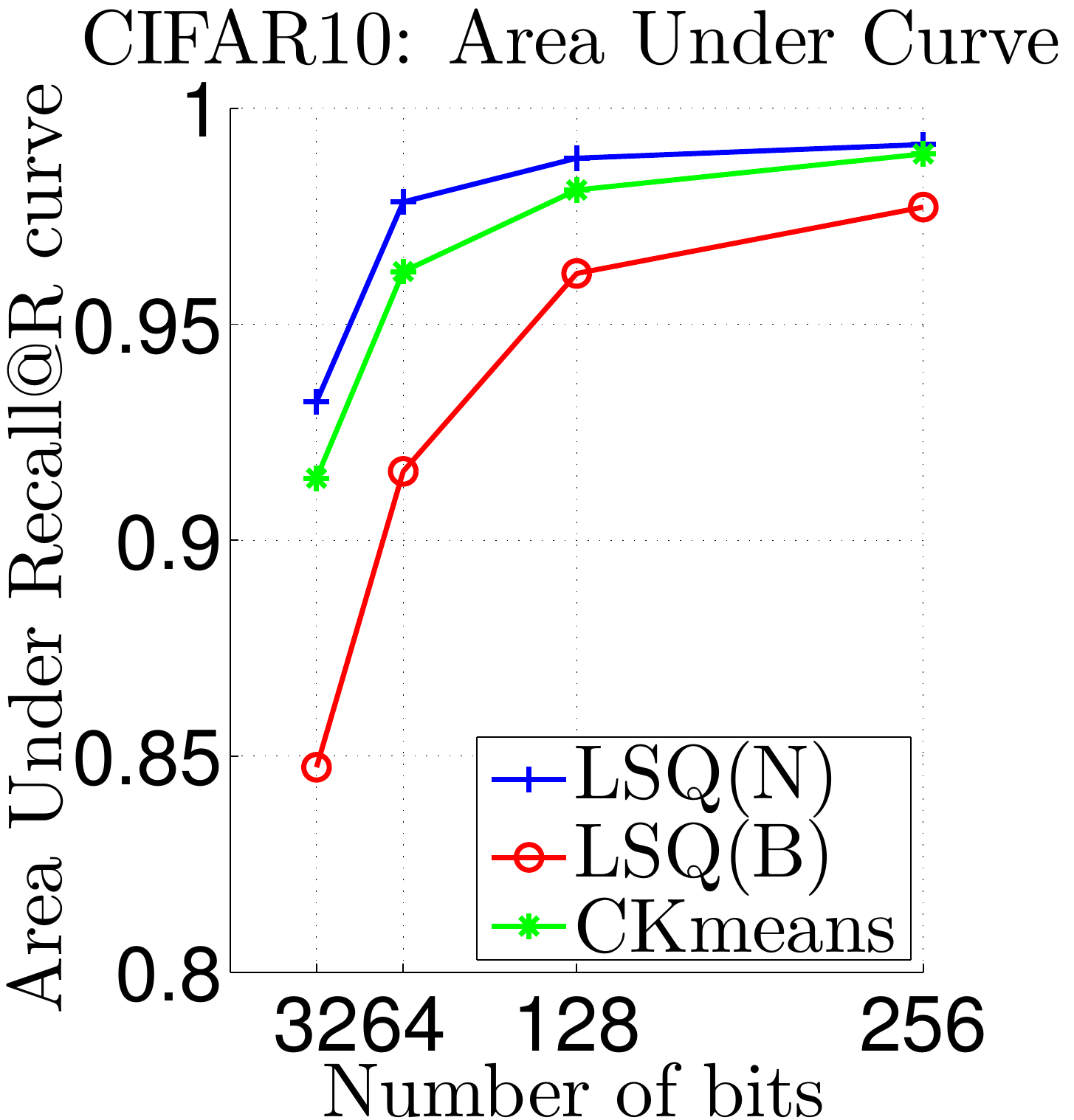}}
\subfloat[]{\includegraphics[width=0.25\textwidth]{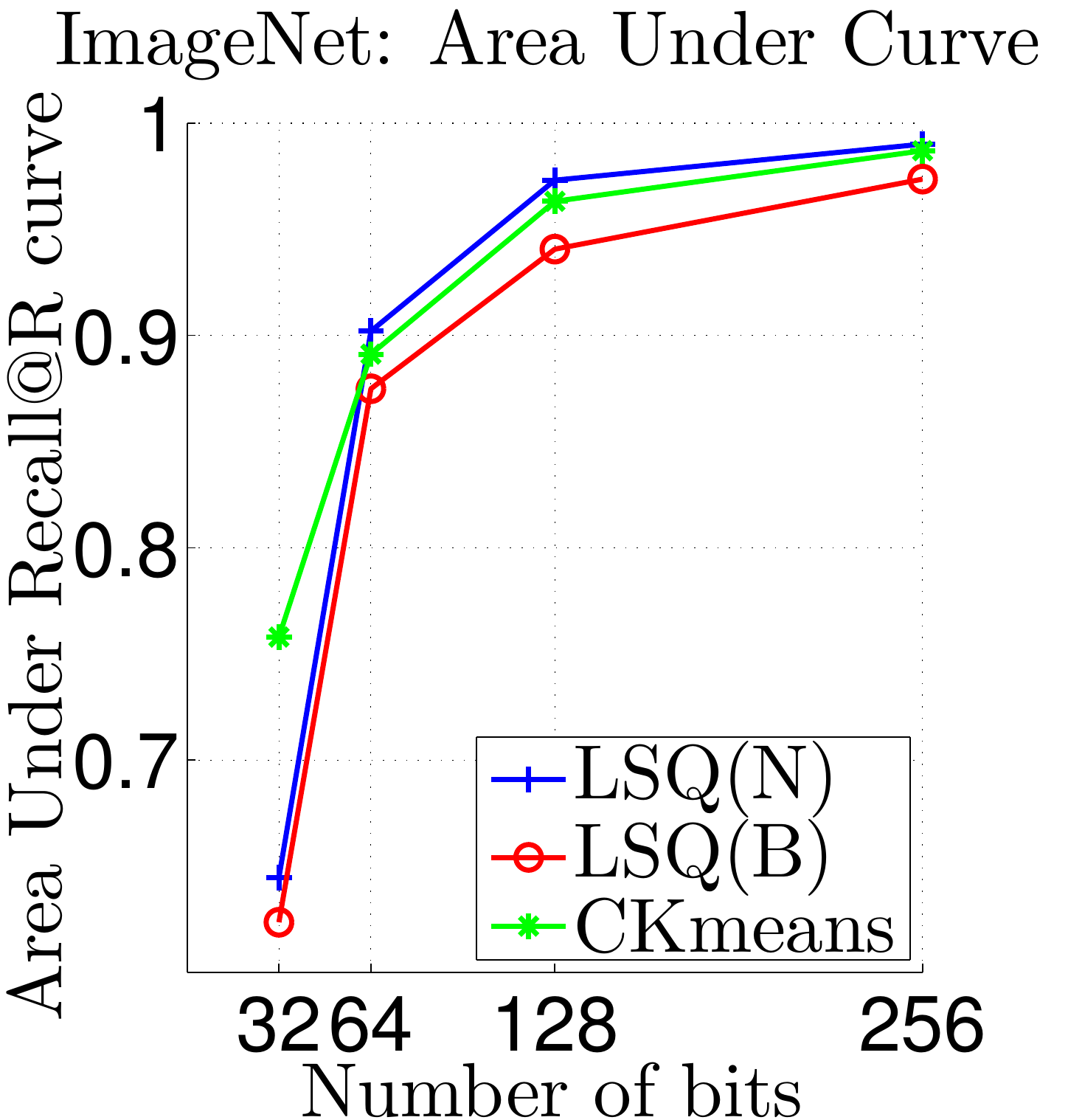}}
\caption{\footnotesize The Area Under Recall@R curves for retrieval using 5 bits per code dimension ( i.e. 32 quantization steps). Each diagram shows the curve for different methods and different amount of total bit budget. (a) Results on CIFAR10 dataset. (b) Results on ImageNet dataset.}
\label{fig:distance_estimation_AUC}
\end{figure}

Figure \ref{fig:distance_estimation_AUC} explores the effect of the number of bits on the different methods. We fixed the number of bits per code dimension to 5 (e.g. the CK-means algorithm would learn 32 clusters per segment) and report the area under the Recall@R curve. Again, LSQ performs better than CK-means.

\subsection{Retrieval using Subset Indexing}
As discussed in section \ref{sec:ret_subset}, either binary or $n$-ary coding can be used to speed up search with Subset Indexing. This approach limits the search to a small number of samples by indexing subsets of the database (subset indexing). Here, the performance of binary and n-ary coding is compared. We compare the retrieval results of the best $n$-ary encoding for this task(CK-means) to the best binary coding(the binary version of LSQ). 

\begin{figure*}
\centering
\subfloat[]{\includegraphics[height=0.26\textwidth]{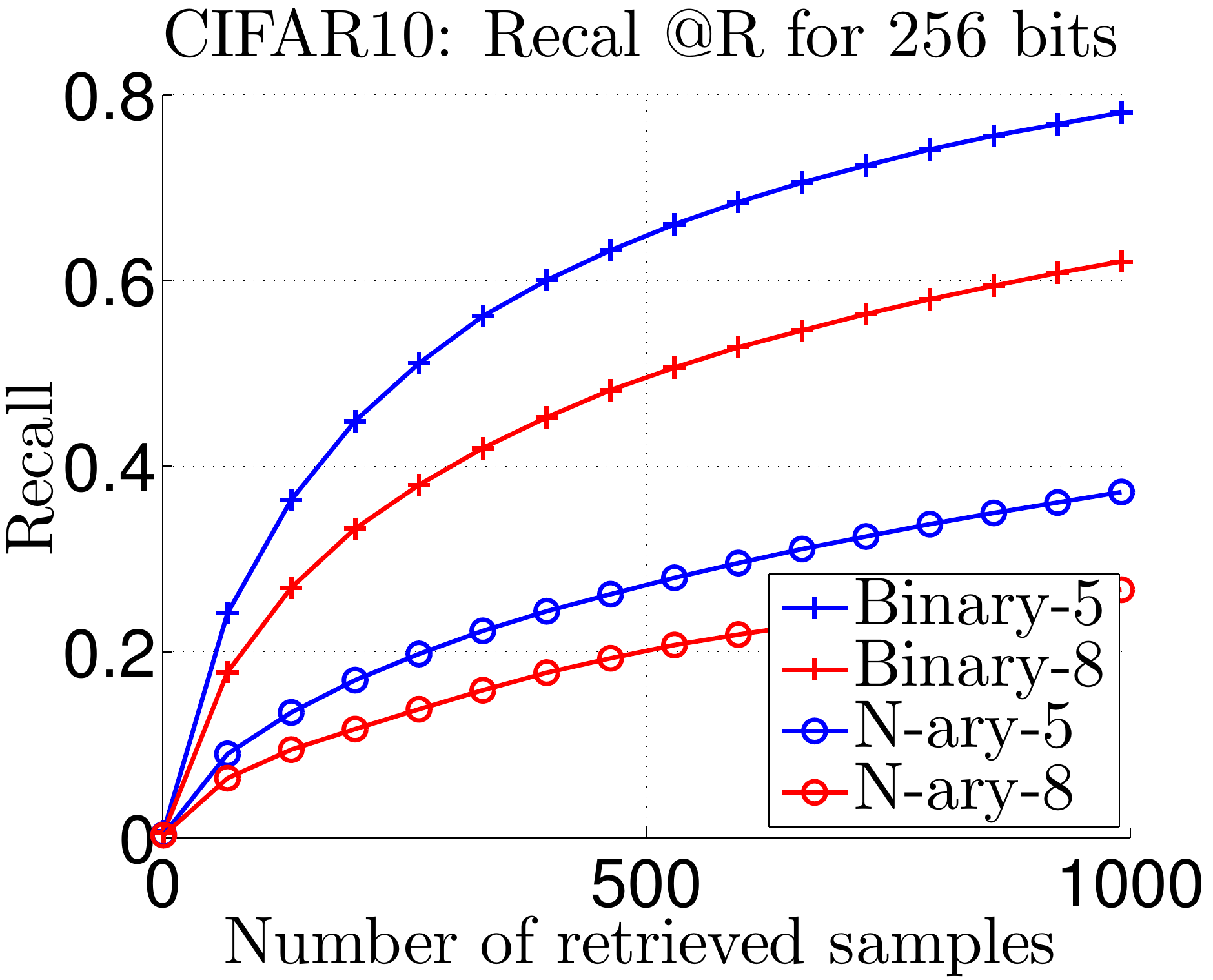}}
\subfloat[]{\includegraphics[height=0.26\textwidth]{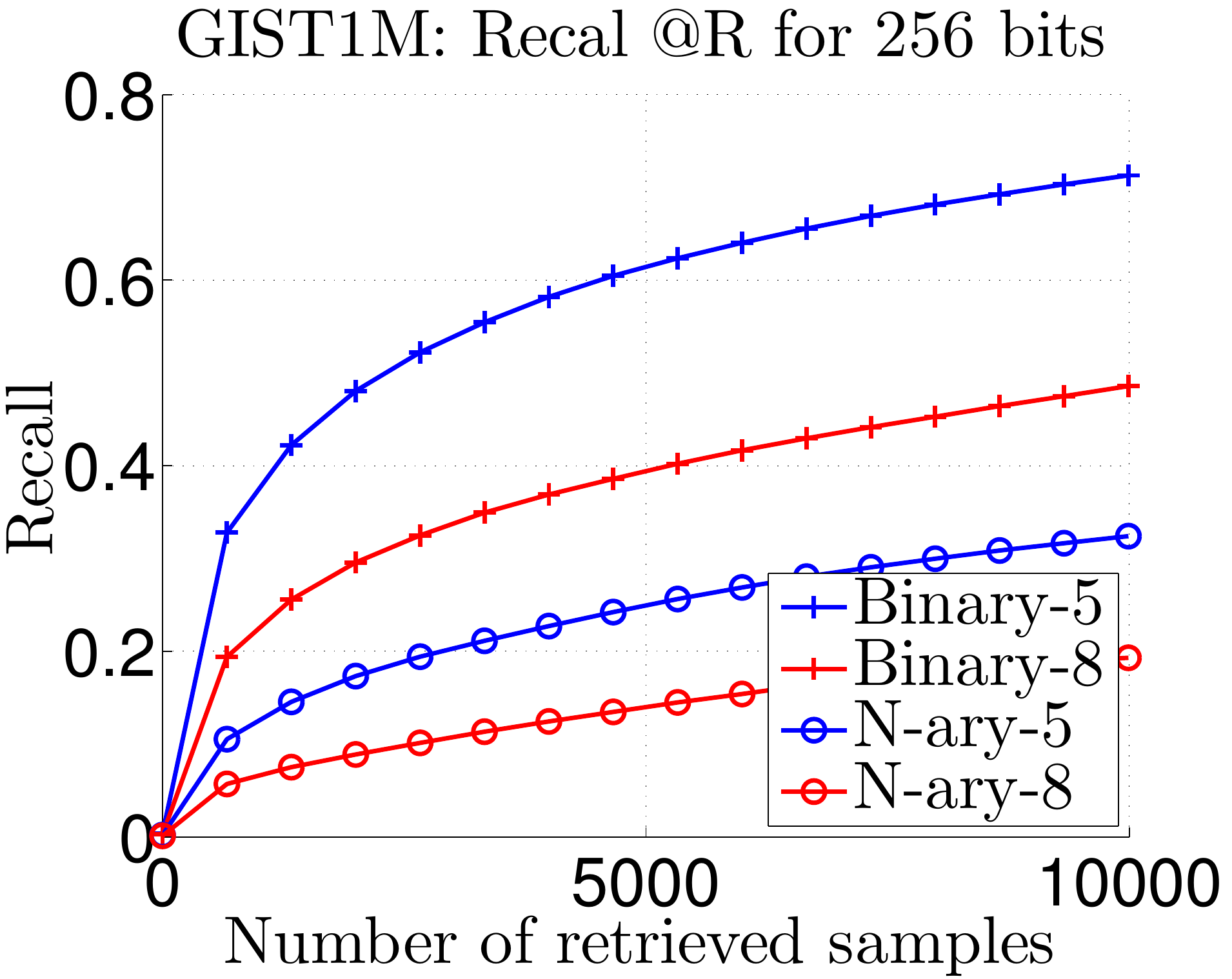}}
\subfloat[]{\includegraphics[height=0.26\textwidth]{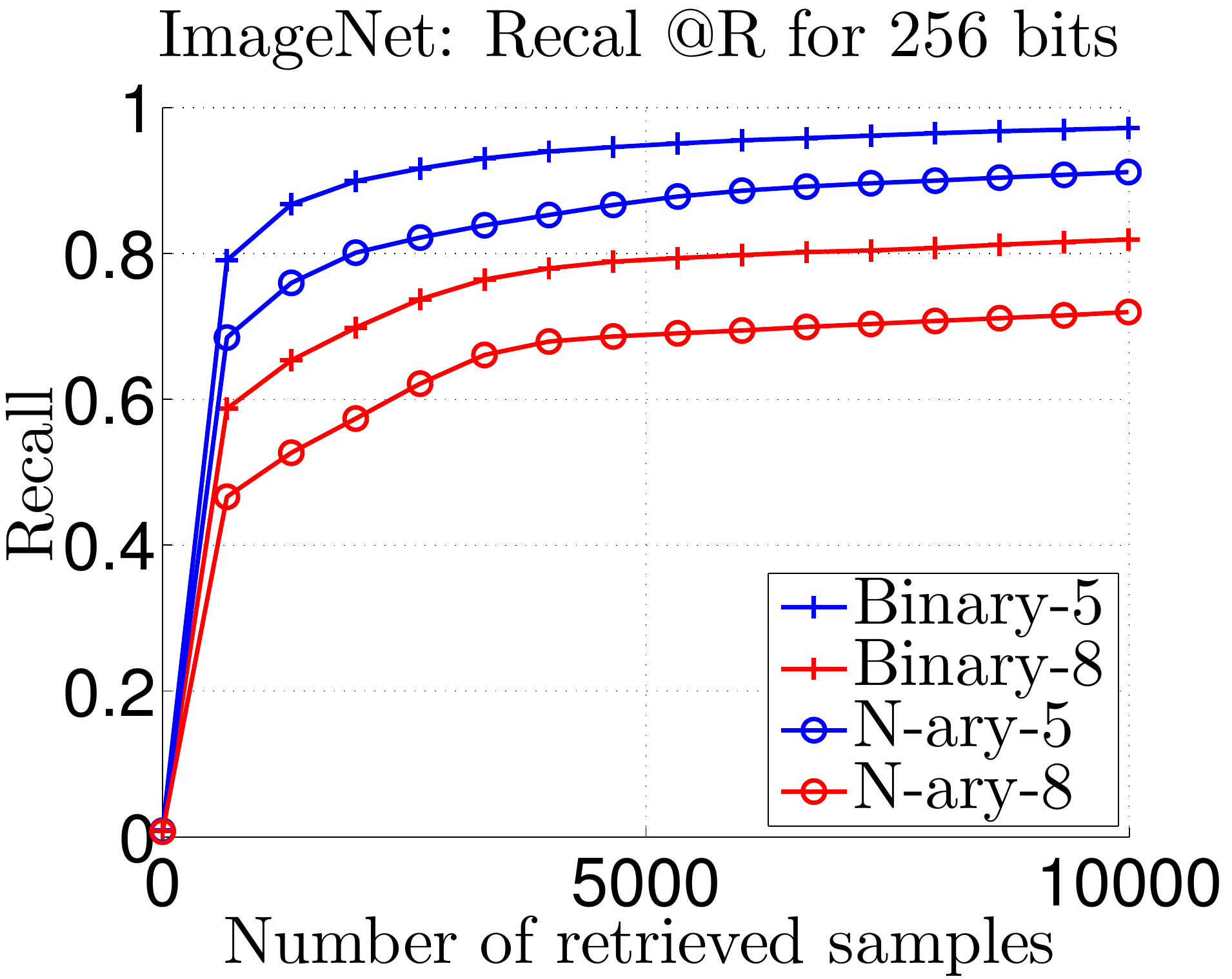}}
\caption{\footnotesize The Recall@R curves for retrieval using subset indexing for 256 bits. Each diagram shows the curve for the best n-ary and binary coding method in this task (CKmeans and Binary LSQ respectively). (a) Results on CIFAR10 dataset. (b) Results  on GIST1M dataset. (c) Results on ImageNet dataset.}
\label{fig:recall_subset_indexing}
\end{figure*}

\begin{figure}
\centering
\subfloat[]{\includegraphics[width=0.17\textwidth]{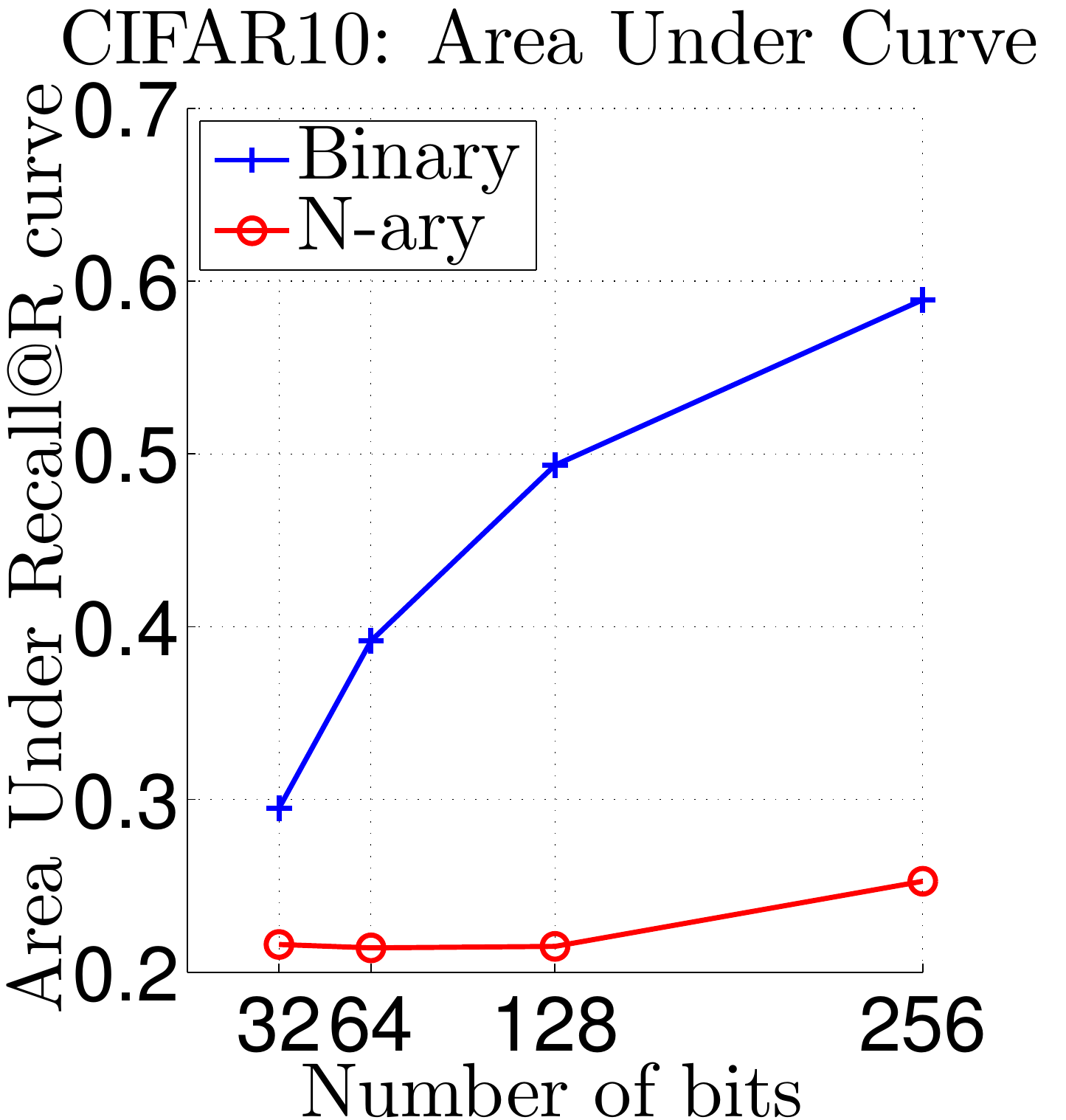}}
\subfloat[]{\includegraphics[width=0.17\textwidth]{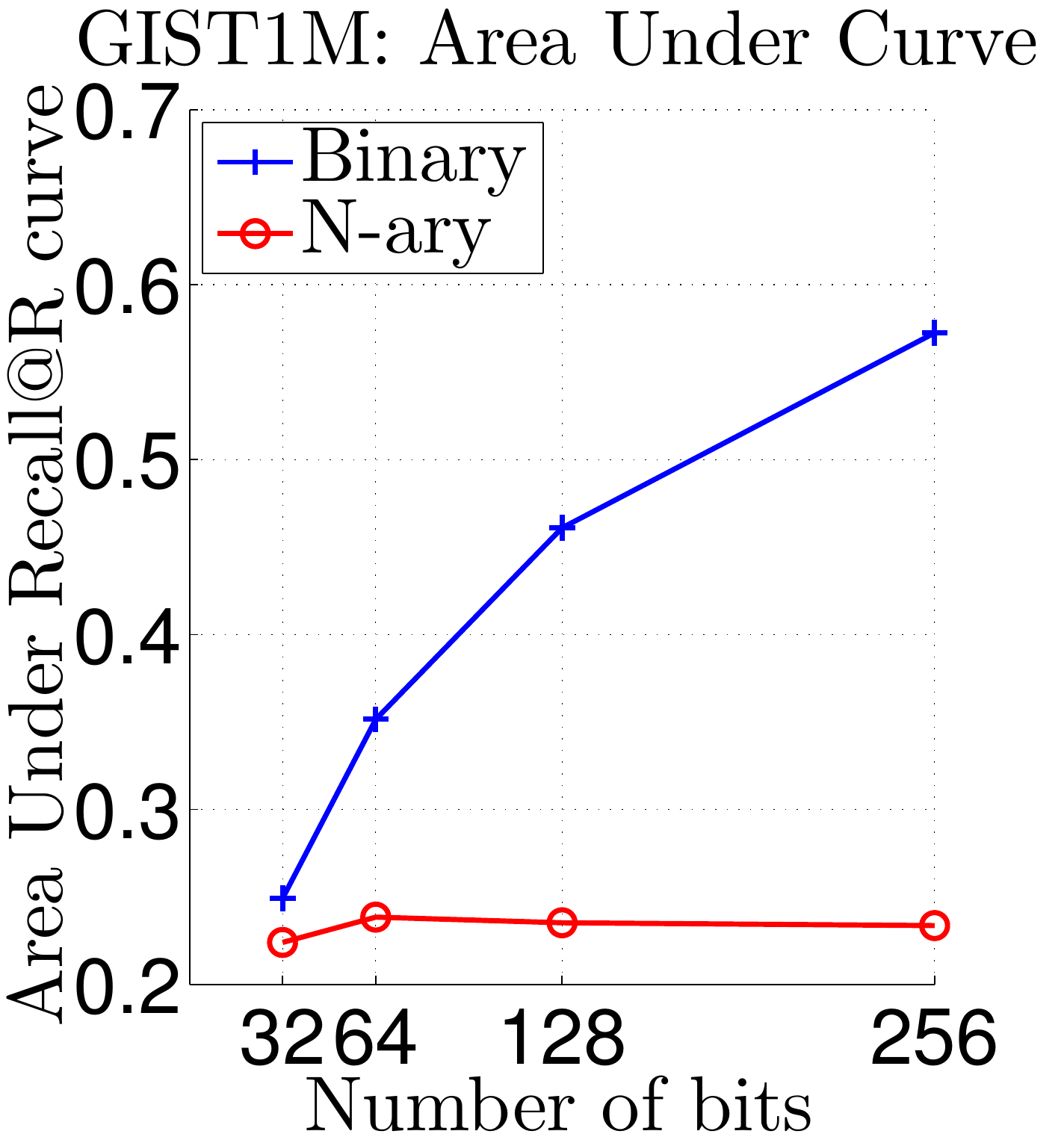}}
\subfloat[]{\includegraphics[width=0.17\textwidth]{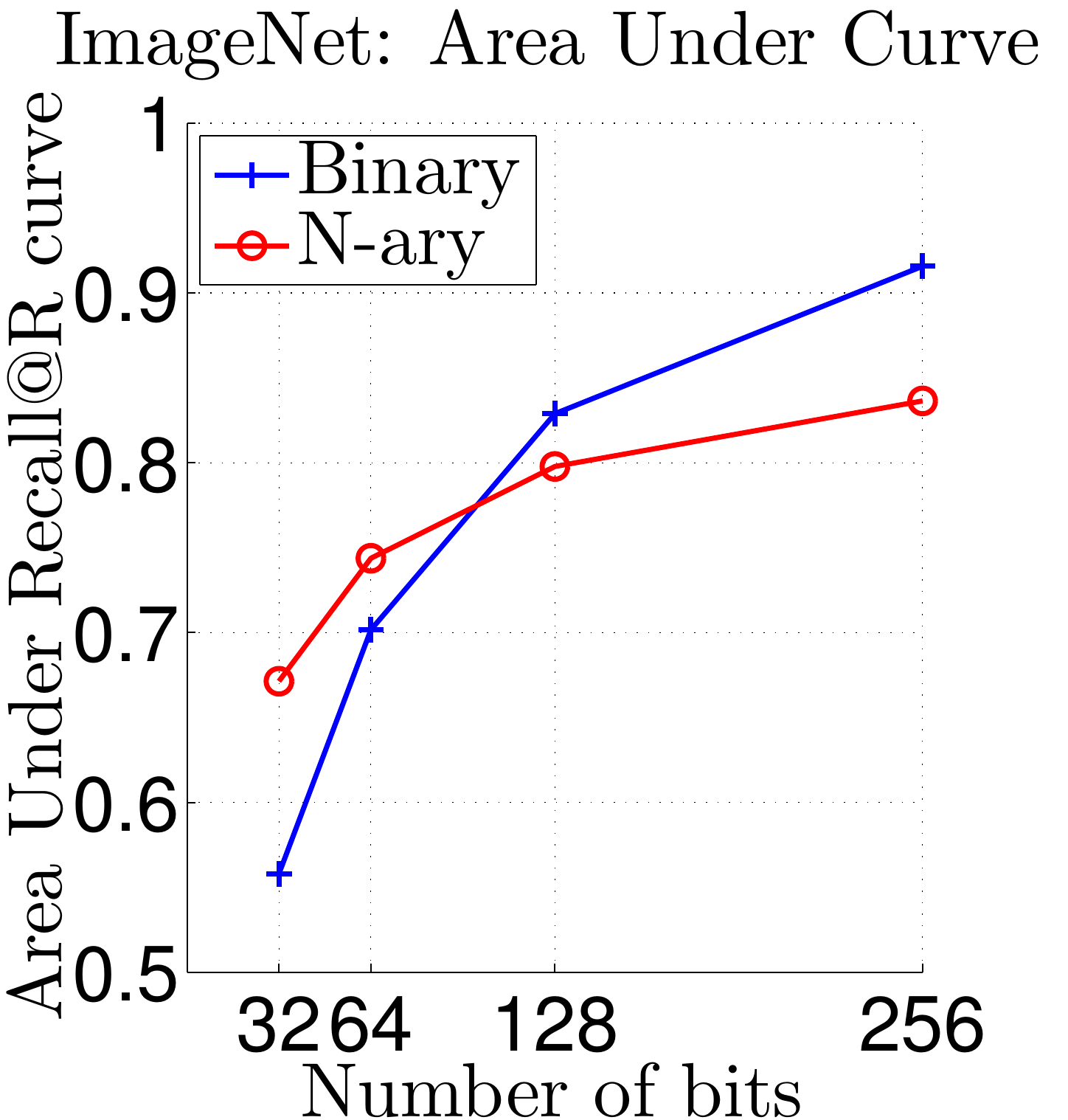}}
\caption{\footnotesize The Area Under Recall@R curves for retrieval using 5 bits per code dimension. Each diagram shows the curve for the best n-ary and binary coding method for this task and different bit budgets. (a) Results on CIFAR10 dataset. (b) Results  on GIST1M dataset. (c) Results on ImageNet dataset.}
\label{fig:auc_subset_indexing}
\end{figure}

Figure \ref{fig:recall_subset_indexing}, shows the recall@R curves for this experiment with varying numbers of bits per code dimension for a fixed budget of 256 bits. In \textit{N-ary-k}, $k$ bits are used for quantizing each dimension and additionally indexing in the multi-index hashing method(i.e. $2^k$ quantization steps for each dimension). Similarly, in \textit{Binary-k}, $k$ consecutive bits are used for indexing in the hashing method. The effect of changing the budget of the encoder on the retrieval task can be seen in Figure \ref{fig:auc_subset_indexing}. These figures illustrate that the binary encoding techniques outperform the n-ary encoders when the subset indexing technique is used, as discussed in Section \ref{sec:ret_subset}.

\textbf{Discussion:} These experiments confirm that when retrieval is performed by \textbf{distance estimation}, it is better to use $n$-ary coding with $n>2$  based on subspace reduction (e.g. LSQ). On the other hand, when \textbf{subset indexing} is used for retrieval, binary coding outperforms n-ary coding.    

\subsection{Comparison of binary coding methods}
\begin{figure}
\centering
\subfloat[]{\includegraphics[width=0.25\textwidth]{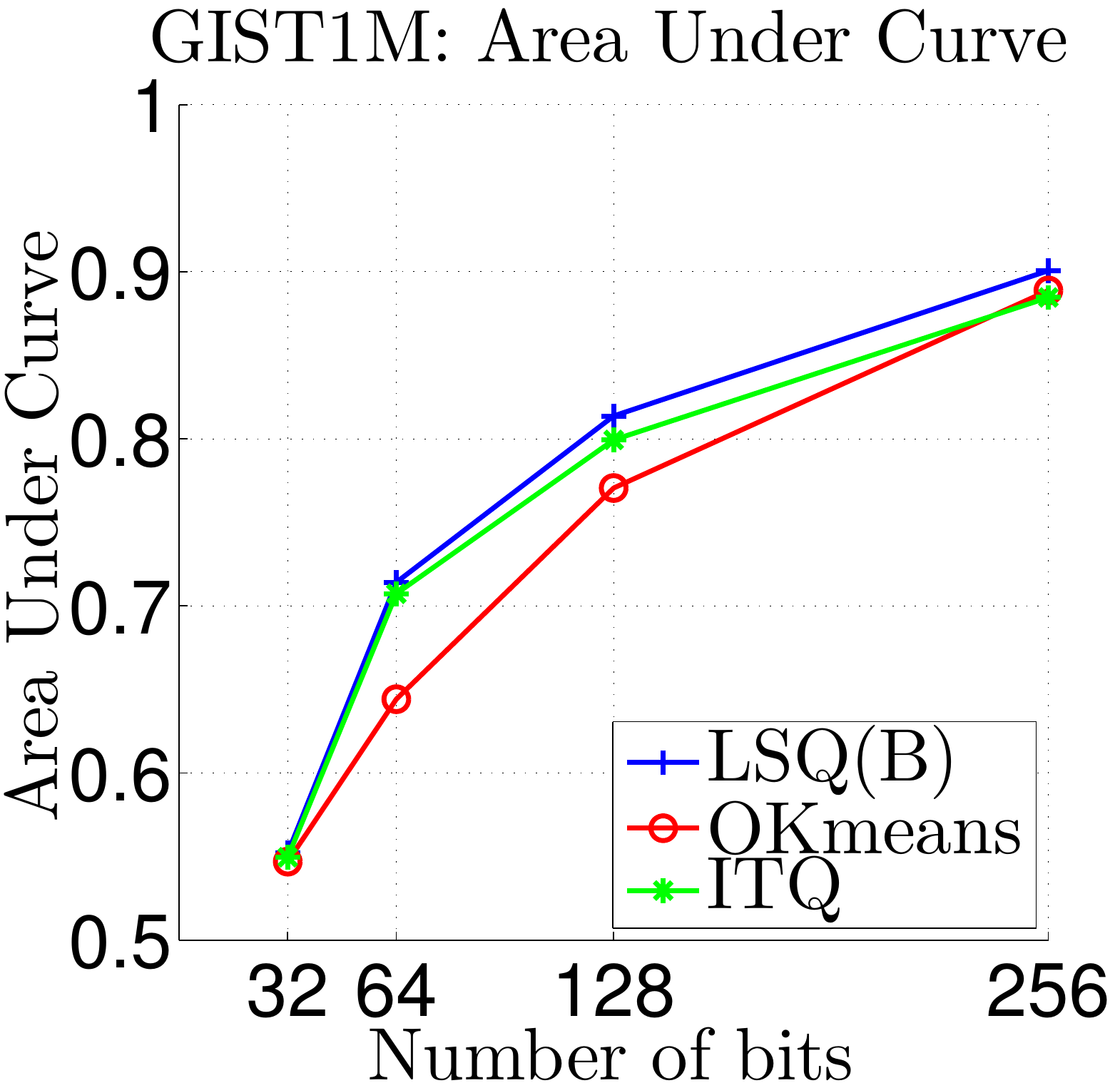}}
\subfloat[]{\includegraphics[width=0.25\textwidth]{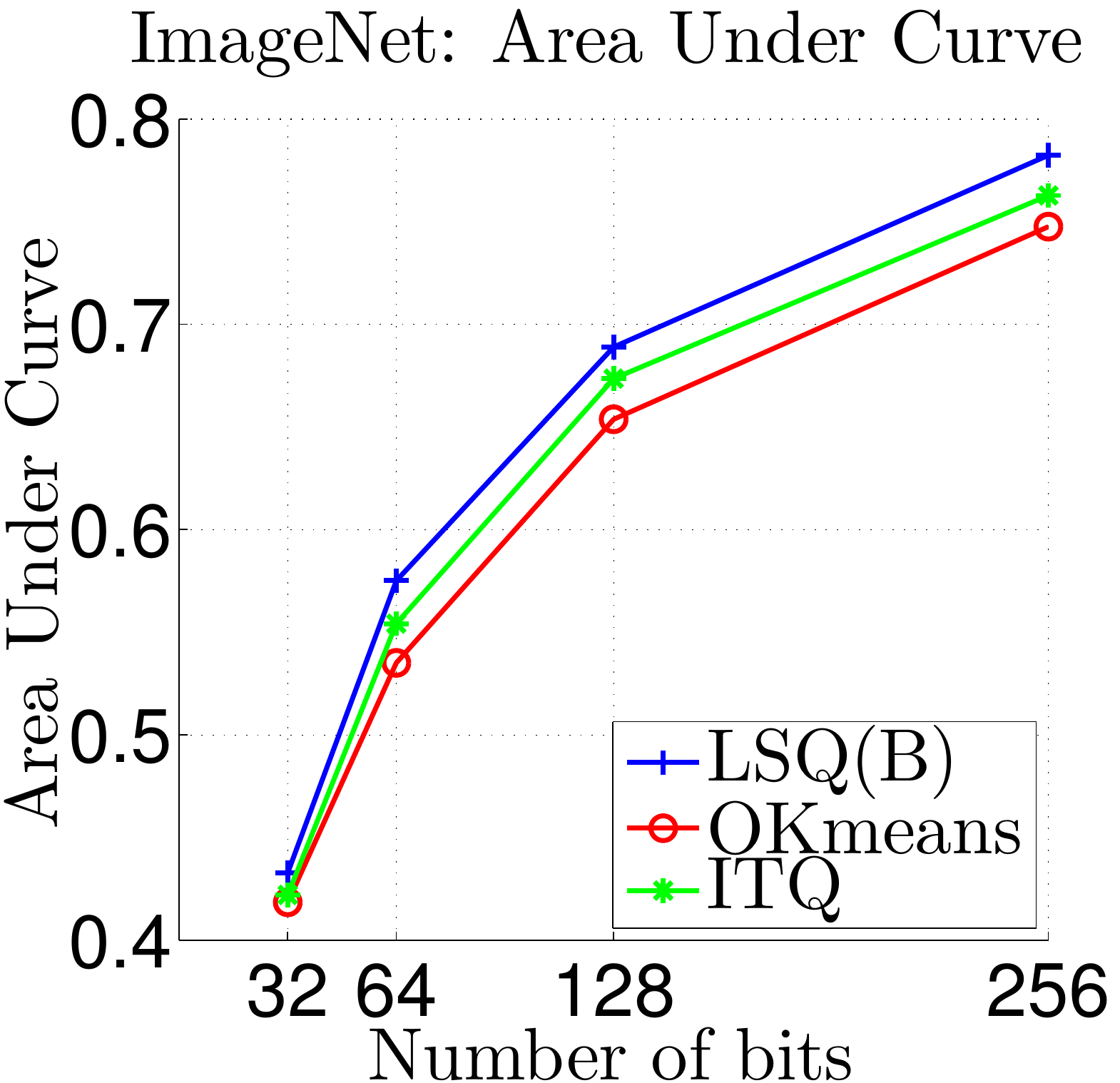}}
\caption{\footnotesize The Area Under precision recall curves for binary methods. Each diagram shows the curve for different methods and different bit budgets. (a) Results on GIST1M dataset. (b) Results  on ImageNet dataset.}
\label{fig:binary_comparision}
\end{figure}

Both CK-means and LSQ can be viewed as generalizations of binary encoding where the number of quantization steps can be more than two. Here, the number of quantization steps is set to two and the binary versions of LSQ and CK-means (namely LSQ(B) and OK-means respectively) are compared with ITQ using subset indexing. Figure \ref{fig:binary_comparision}, shows the area under the recall precision curve for these three binary coding methods under varying bit budgets. As can be seen, the binary version of LSQ outperforms ITQ and the binary version of CK-means.

\subsection{Convergence of the algorithms}
\begin{figure}
\centering
\subfloat[]{\includegraphics[width=0.155\textwidth,height=0.14\textwidth, natwidth = 6.79in, natheight = 6.12in]{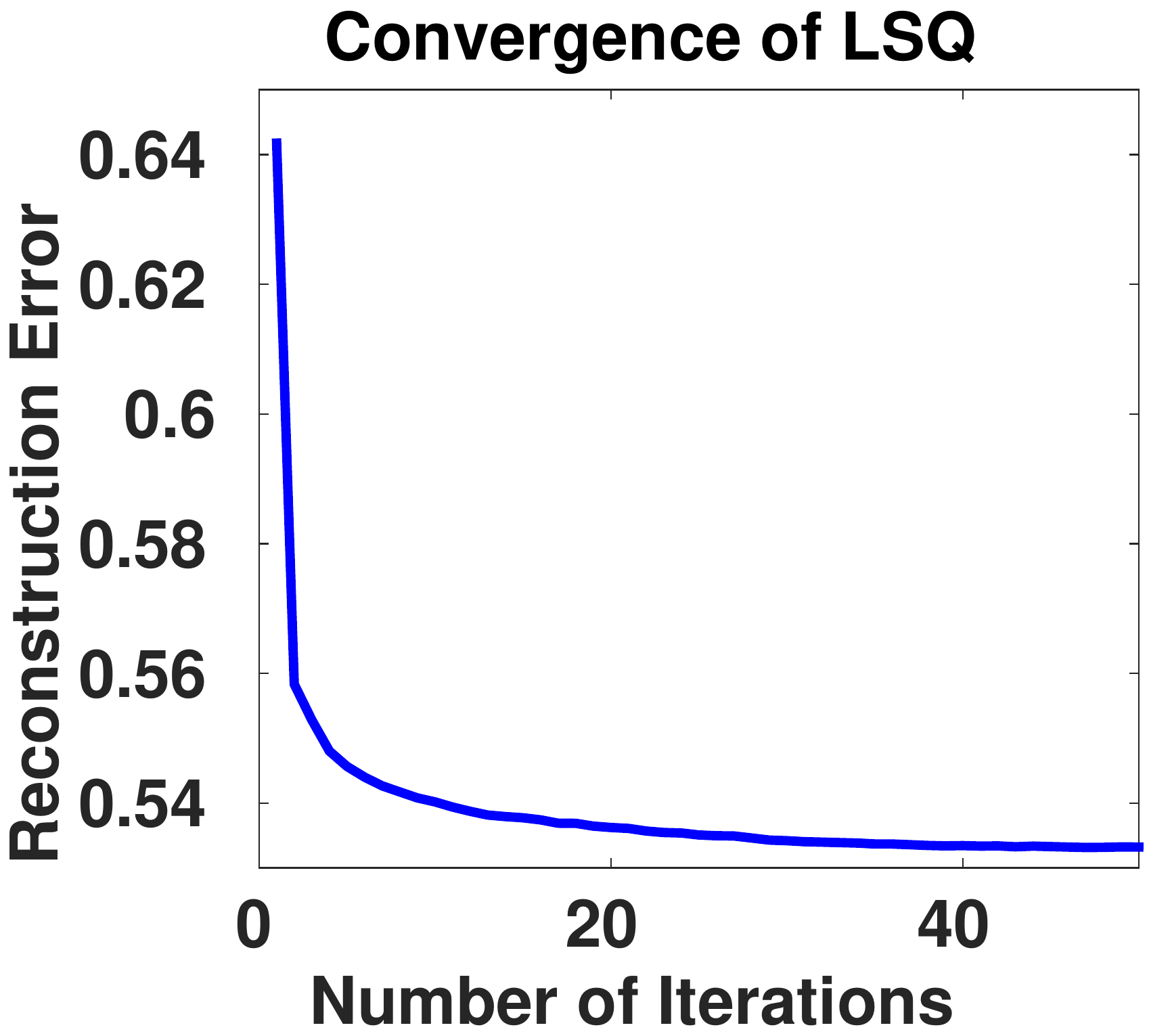}}
\subfloat[]{\includegraphics[width=0.155\textwidth,height=0.14\textwidth, natwidth = 6.8in, natheight = 5.88in]{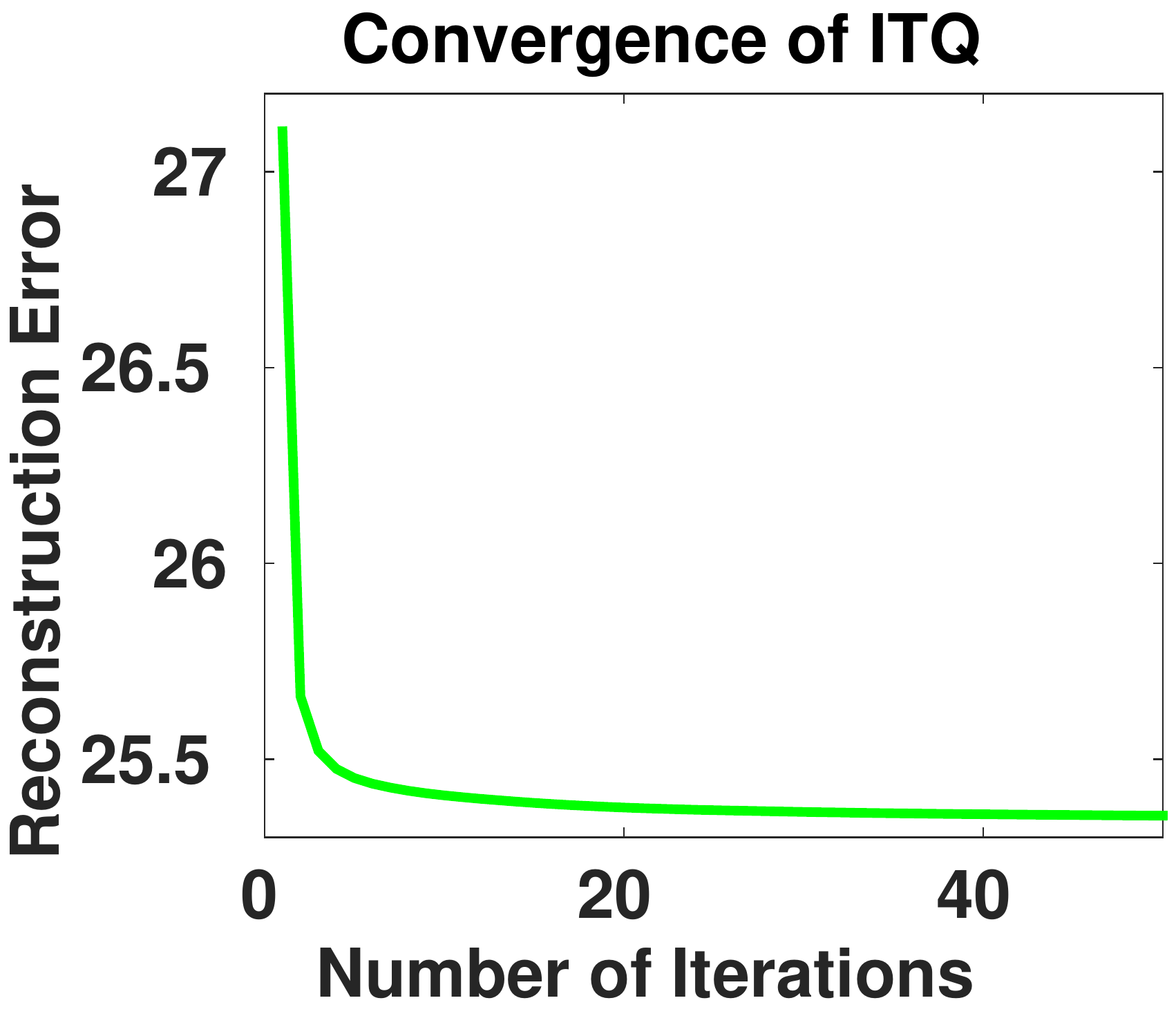}}
\subfloat[]{\includegraphics[width=0.165\textwidth,height=0.14\textwidth, natwidth = 6.82in, natheight = 5.84in]{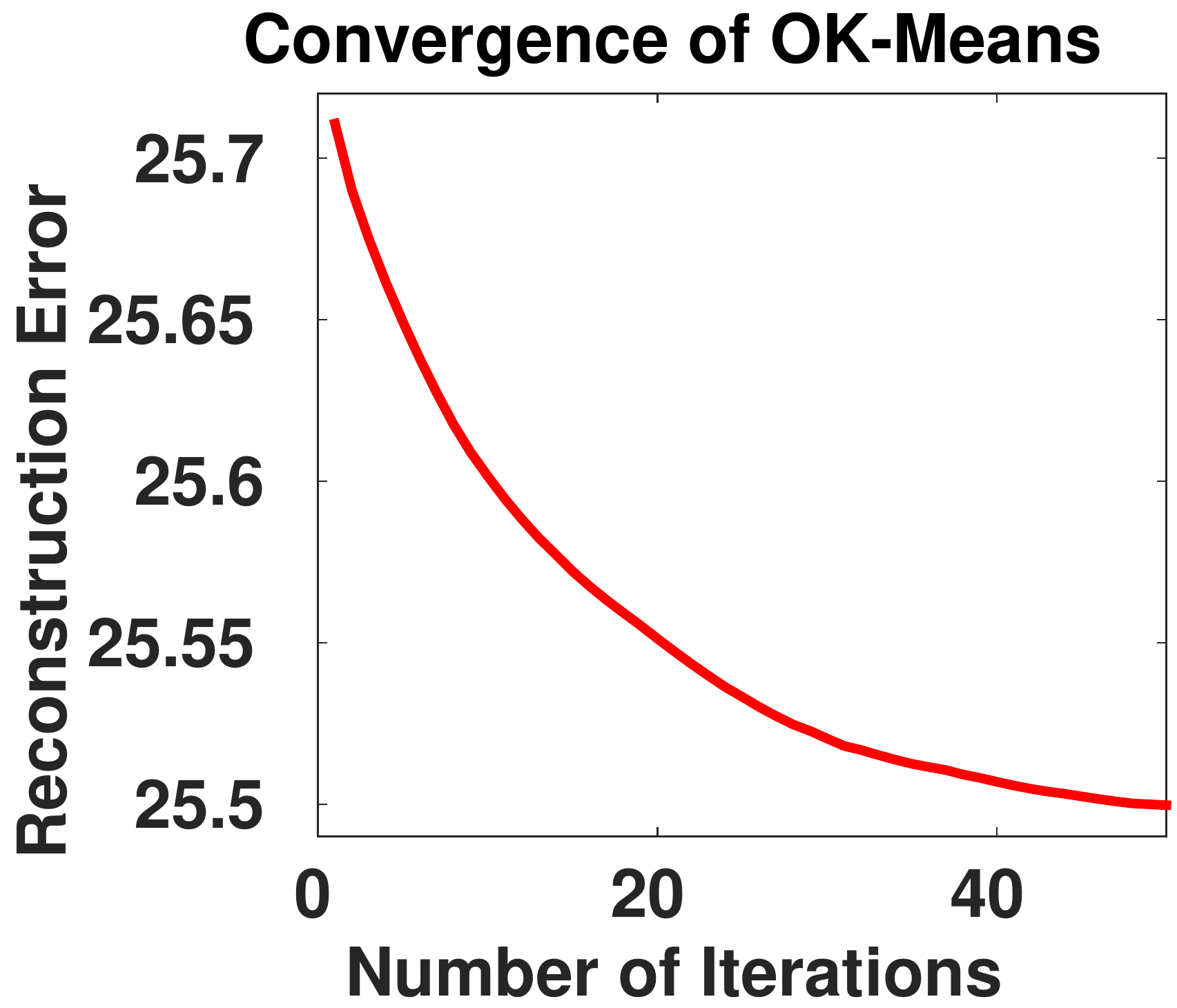}}
\caption{\footnotesize The convergence of different coding algorithms. (a) LSQ(B) (b) ITQ (c) OK-Means. Note different scale of reconstruction error are required.}
\label{fig:convergence}
\end{figure}

In Figure \ref{fig:convergence}, the convergence of different binary coding methods are shown. For this experiment, GIST1M is used. As can be seen, LSQ converges much faster than OK-Means. Also note that the final reconstruction error of LSQ is much smaller than ITQ and OK-means, reflecting the fact that LSQ reconstructs the data more accurately using the same memory budget.

\subsection{$n$-ary Codes as Feature Vectors}
\label{sec:exp_embedding}
\begin{figure}
\centering
\includegraphics[width=0.3\textwidth]{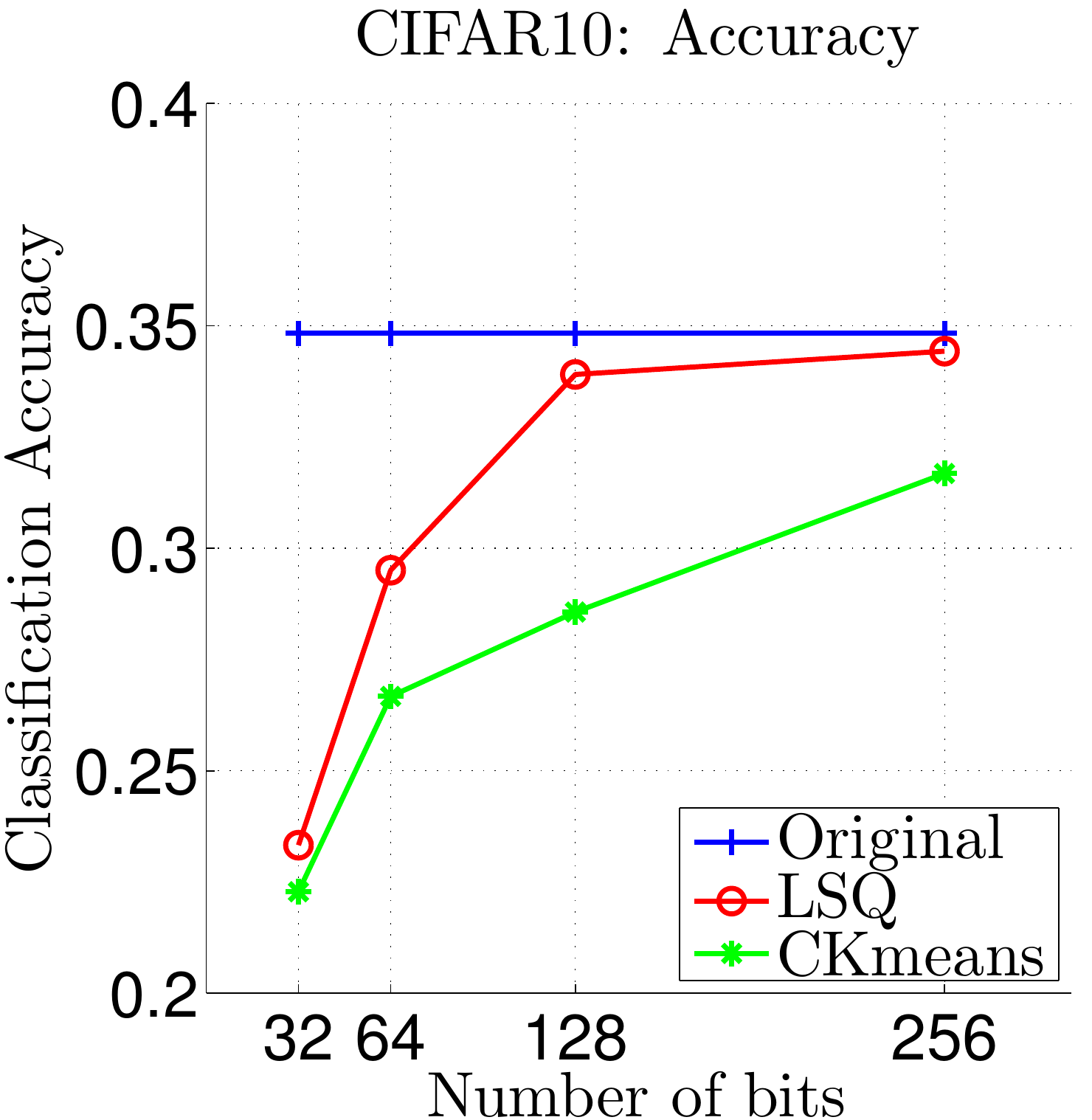}
\caption{\footnotesize Classification accuracy using the codes of different methods as feature vectors.}
\label{fig:embedding}
\end{figure}

The codes constructed by LSQ can be used as feature vectors to perform learning tasks. Figure \ref{fig:embedding} shows the performance evaluation of a classification task using different codings as features. As proposed in sec \ref{sec:embedding}, for CK-means, we refine the index assignments to clusters by mapping the cluster centers in each subspace into a one dimensional space using PCA and convert each dimension of the code to the corresponding value in this 1D space. It can be seen that our proposed quantization method outperforms CK-means even after refining the CK-Means index assignments.

\section{Conclusion}
\label{sec:con}
We focused on the problem of large scale retrieval using ANN. A new general approach for multi-dimensional $n$-ary coding -Linear Subspace Quantization (LSQ)- was introduced for ANN. LSQ achieves lower reconstruction error than other $n$-ary coding methods. Furthermore, it preserve the similarities in the original space, which is important when it is used directly for learning tasks. Experiments show that LSQ outperforms other binary and n-ary coding methods on large scale image retrieval. We also compared the performance of binary and n-ary coding methods for this task. We showed that $n$-ary coding outperforms binary coding when distance estimation is used to reduce the search computation cost. However, in combination with Subset Indexing, interestingly, binary coding works better for retrieval.

\pagebreak
{\small
\bibliographystyle{ieee}
\bibliography{main}
}

\end{document}